\documentclass[journal]{IEEEtran}
\usepackage[utf8]{inputenc}
\usepackage[english]{babel}
\usepackage{makeidx}
\usepackage{graphicx}
\usepackage[hidelinks]{hyperref}
\usepackage{amsmath}
\usepackage[inline]{enumitem}
\usepackage{multirow}
\usepackage{subcaption}
\usepackage{caption}
\usepackage{color}
\usepackage{booktabs}
\usepackage{siunitx}
\usepackage{cite}
\usepackage{algpseudocode}
\usepackage{algorithmicx}
\usepackage{array}
\usepackage{float}
\usepackage{todonotes}
\usepackage[keeplastbox]{flushend}
\usepackage[linesnumbered,ruled,vlined]{algorithm2e}
\usepackage{amsfonts}
\usepackage{threeparttable}

\algnewcommand\algorithmicforeach{\textbf{for each}}
\algdef{S}[FOR]{ForEach}[1]{\algorithmicforeach\ #1\ \algorithmicdo}

\begin{document}
%
\title{Cascaded Structure Tensor Framework for Robust Identification of Heavily Occluded Baggage Items from X-ray Scans}
%
%
%

\author{Taimur~Hassan\textsuperscript{*}, 
        Samet~Ak\c{c}ay, 
        Mohammed~Bennamoun,~\IEEEmembership{Senior Member,~IEEE,}
        Salman~Khan,
        and~Naoufel~Werghi,~\IEEEmembership{Senior~Member,~IEEE}
\thanks{This work is supported with a research fund from Khalifa University: Ref: CIRA-2019-047.}
\thanks{T. Hassan and N. Werghi are with the Center for Cyber-Physical Systems (C2PS), Department of Electrical and Computer Engineering, Khalifa University, Abu Dhabi, UAE.}
\thanks{S. Ak\c{c}ay is with the Department of Computer Science Durham University, and COSMONiO AI, Durham UK.}
\thanks{M. Bennamoun is with the Department of Computer Science and Software Engineering, The University of Western Australia, Perth Australia.}
\thanks{S. Khan is with the Inception Institute of Artificial Intelligence, Abu Dhabi, UAE.}
\thanks{* Corresponding Author. Email: taimur.hassan@ku.ac.ae}}

%

\maketitle

\begin{abstract}
 In the last two decades, baggage scanning has globally become one of the prime aviation security concerns. Manual screening of the baggage items is tedious, error-prone, and compromise privacy. Hence, many researchers have developed X-ray imagery-based autonomous systems to address these shortcomings. This paper presents a cascaded structure tensor framework that can automatically extract and recognize suspicious items in heavily occluded and cluttered baggage. The proposed framework is unique, as it intelligently extracts each object by iteratively picking contour-based transitional information from different orientations and uses only a single feed-forward convolutional neural network for the recognition. The proposed framework has been rigorously evaluated using a total of 1,067,381 X-ray scans from publicly available GDXray and SIXray datasets where it outperformed the state-of-the-art solutions by achieving the mean average precision score of 0.9343 on GDXray and 0.9595 on SIXray for recognizing the highly cluttered and overlapping suspicious items. Furthermore, the proposed framework computationally achieves 4.76\% superior run-time performance as compared to the existing solutions based on publicly available object detectors.

\end{abstract}


\IEEEpeerreviewmaketitle

\section{Introduction}
\IEEEPARstart{X}{-ray} imaging is a widely adopted tool for the baggage inspection at airports, malls and cargo transmission trucks \cite{_3}. Baggage threats have become a prime concern all over the world. According to a recent report, approximately 1.5 million passengers are searched every day in the United States against weapons and other dangerous items \cite{_4}. 
Manual screening process is resource-intensive and requires constant attention of human experts during monitoring. This introduces an additional risk of human-error caused by fatigued work-schedule, difficulty in catching contraband items, requirement of quick decision making, or simply due to a less experienced operator.
\begin{figure}[htb]
\includegraphics[scale=0.1065]{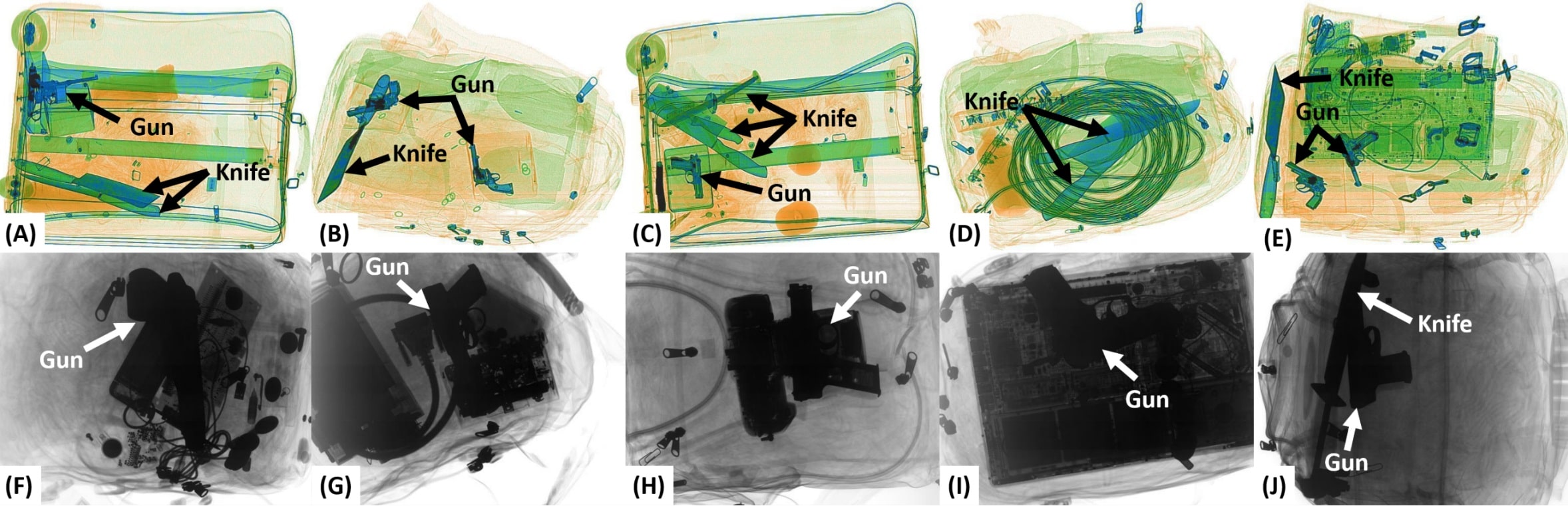}
\caption{\small Exemplar X-ray images showing heavily occluded and cluttered items. Top row shows scans from SIXray dataset \cite{_46} while bottom row shows scans from GDXray dataset \cite{_55}. }
\centering
\label{fig:block1}
\end{figure}
Therefore, aviation authorities, all over the world, are actively looking for automated and reliable baggage screening systems to increase the inspection speed and support operator alertness. Automated inspection is also desirable for privacy concerns. 
A number of large-scale natural image datasets for object detection are publicly available, enabling the development of popular object detectors like R-CNN \cite{_8}, SPP-Net \cite{_9}, YOLO \cite{_10} and RetinaNet \cite{_11}.
In contrast, only a few datasets for X-ray images are currently available for researchers to develop robust computer-aided screening systems. Also, the nature of radiographs is quite different than natural photographs. Although they can reveal the information invisible in the normal photographs (due to radiations), they lack texture (especially the grayscale X-ray scans), due to which conventional detection methods do not work well on them \cite{_12}. In general, screening objects and anomalies from baggage X-ray (grayscale or colored) scans is a challenging task, especially when the objects are closely packed to each other, leading to heavy occlusion. In addition to this, baggage screening systems face a severe class imbalance problem due to the low suspicious to normal items ratio. Therefore, it is highly challenging to develop an unbiased decision support system which can effectively screen baggage items because of  the high contribution of the normal items within the training images. Figure \ref{fig:block1} shows some of the X-ray baggage scans where the suspicious items such as \textit{guns} and \textit{knives} are highlighted in a heavily occluded and cluttered environment. 

\section{Related Work}
\noindent Several methods for detecting suspicious objects in X-ray imagery have been proposed in the literature. These can be categorized as conventional machine learning methods and deep learning methods. We provide a representative list of the main approaches, and we refer the reader to the work of \cite{ackay2020} for an exhaustive survey.

\subsection{Traditional Approaches}
\noindent Many researchers have used traditional machine learning (ML) methods to recognize baggage items from the X-ray scans. Bastan et al. \cite{_13} proposed a structured learning framework that encompasses dual-energy levels for the computation of low textured key-points to detect \textit{laptops}, \textit{handguns} and \textit{glass bottles} from multi-view X-ray imaging. They also presented a framework that utilizes Harris, SIRF, SURF and FAST descriptors to extract key-points which are then passed to a bag of words (BoW) model for the classification and retrieval of X-ray images \cite{_14}. They concluded that although BoW produces promising results on regular images, it does not perform well on low textured X-ray images \cite{_14}. Jaccard et al. \cite{_35} proposed an automated method for detecting \textit{cars} from the X-ray cargo transmission images based upon their intensity, structural and symmetrical properties using a random forest classifier. Turcsany et al. \cite{_38} proposed a framework that uses the SURF descriptor to extract distinct features and passes them to a BoW for object recognition in baggage X-ray images. Riffo et al. \cite{_39} proposed an adapted implicit shape model (AISM) that first generates a category-specific codebook for different parts of targeted objects along with their spatial occurrences (from training images) using SIFT descriptor and agglomerative clustering. Then, based upon voting space and matched codebook entries, the framework detects the target object from the candidate test image. 

\begin{figure*}[htb]
\includegraphics[scale=0.135]{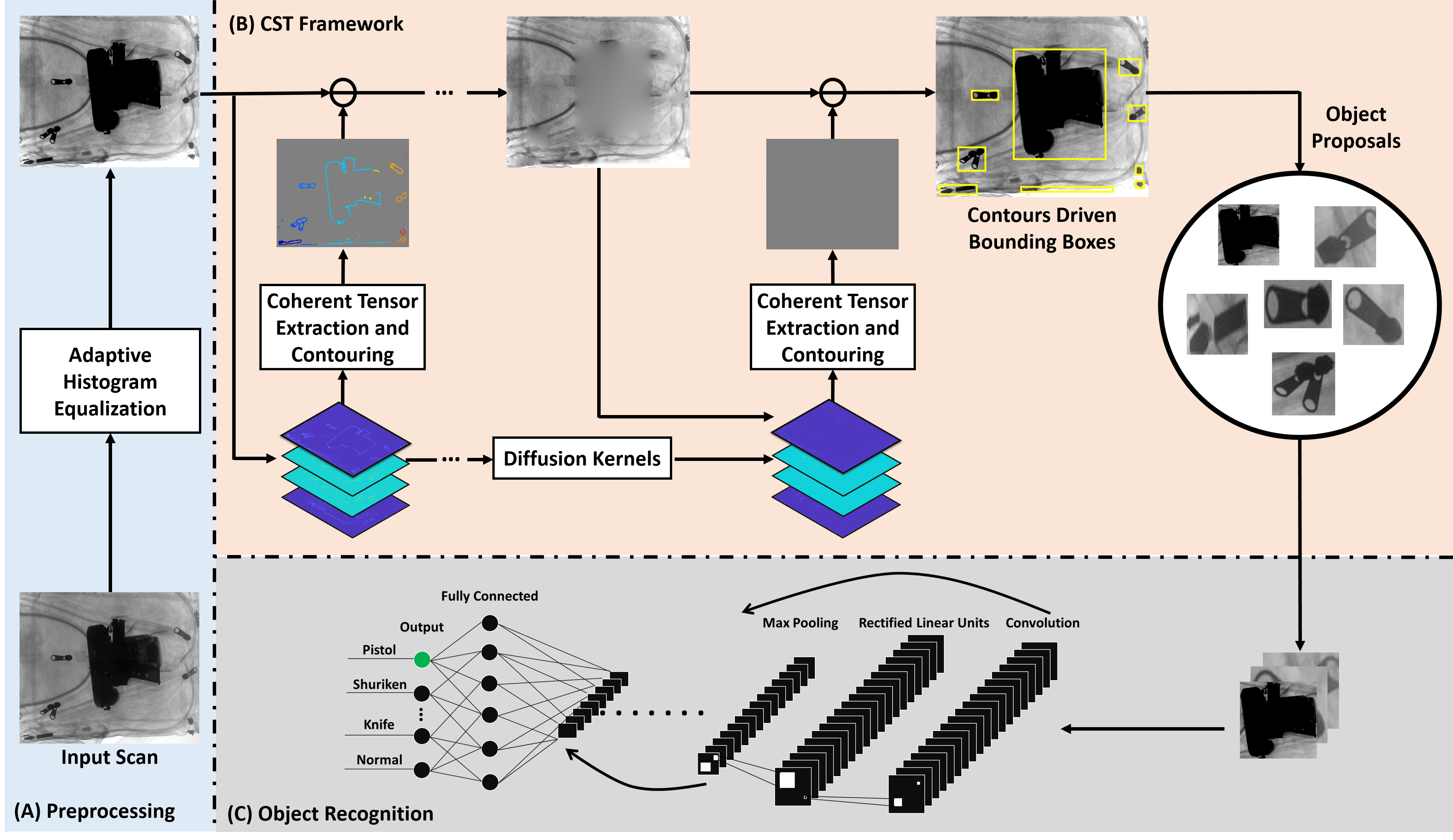}
\caption{\small Block diagram of the proposed framework. The input scan is first preprocessed (A). Afterwards, the proposal for each baggage item is automatically extracted through the CST framework (B). The extracted proposals are then passed to the pre-trained model for recognition (C).}
\centering
\label{fig:block2}
\end{figure*}
\subsection{Deep Learning Based Suspicious Items Detection}
\noindent Recently, many researchers have presented work employing deep learning architectures for the detection and classification of suspicious baggage items from X-ray imagery. These studies are either focused on the usage of supervised classification models or unsupervised adversarial learning:
\subsubsection{Unsupervised Anomaly Detection}
\noindent{Ak\c{c}ay et al. proposed GANomaly \cite{_27} and Skip-GANomaly \cite{_28} architectures to detect different anomalies from X-ray scans. These approaches employ an encoder-decoder for deriving a latent space representation used by discriminator network to classify anomalies. Both architectures are trained on normal distributions while they are tested on normal and abnormal distributions from CIFAR-10, Full Firearm vs Operational Benign (FFOB) and the local in-house datasets (GANomaly is also verified on MNIST dataset).} 
\subsubsection{Supervised Approaches}
\noindent Ak\c{c}ay et al. \cite{_30} proposed using a pre-trained GoogleNet model \cite{googlenet} for object classification from X-ray baggage scans. They prepared their in-house dataset and tested their proposed framework to detect \textit{cameras}, \textit{laptops}, \textit{guns}, \textit{gun components} and \textit{knives} (mainly, \textit{ceramic knives}). A subsequent work \cite{_31} compared different frameworks for the object classification from X-ray imagery. They concluded that AlexNet \cite{alexnet} as a feature extractor with support vector machines (SVM) performs better than other ML methods. For occluded data, they compared the performance of sliding window-based CNN (SW-CNN), Faster R-CNN \cite{_63}, region-based fully convolutional networks (R-FCN) \cite{_66}, and YOLOv2 \cite{_62} for object recognition. They used their local datasets as well as non-publicly available FFOB and Full Parts Operation Benign (FPOB) datasets in their experimentations. Likewise, Dhiraj et al. \cite{_42} used YOLOv2 \cite{_62}, Tiny YOLO \cite{_62} and Faster R-CNN  \cite{_63} to extract \textit{guns}, \textit{shuriken}, \textit{razor blades} and \textit{knives} from baggage X-ray scans of the GRIMA X-ray database (GDXray \cite{_55}) where they achieved an accuracy of up to 98.4\%. Furthermore, they reported the computational time of 0.16 seconds to process a single image. Gaus et al. \cite{_29} proposed a dual CNN based framework in which the first CNN model detects the object of interest while the second classifies it as benign or malignant. For object detection, the authors evaluated Faster R-CNN \cite{_63}, Mask R-CNN \cite{_61} and RetinaNet \cite{_11} using ResNet\textsubscript{18} \cite{_48}, ResNet\textsubscript{50} \cite{_48}, ResNet\textsubscript{101} \cite{_48}, SqueezeNet \cite{_64} and VGG-16 \cite{_65} as a backbone.
Recently, Miao et al. \cite{_46} provided one of the most extensive and challenging X-ray imagery dataset (named as SIXray) for detecting suspicious items. This dataset contains 1,059,321 scans with heavily occluded and cluttered objects out of which 8,929 scans contain suspicious items. Furthermore, the dataset anticipates the class imbalance problem in real-world scenarios by providing different subsets in which positive and negative samples ratios are varied \cite{_46}. The authors have also developed a deep class-balanced hierarchical refinement (CHR) framework that iteratively infers the image content through reverse connections and uses a custom class-balanced loss function to accurately recognize the contraband items in the highly imbalanced SIXray dataset. They also evaluated the CHR framework on the ILSVRC2012 large scale image classification dataset. After the release of the SIXray dataset, Gaus et al. \cite{_32} evaluated Faster R-CNN \cite{_63}, Mask R-CNN \cite{_61}, and RetinaNet  \cite{_11} on it, as well as on other non-publicly available datasets.
 
\noindent To the best of our knowledge, all the methods proposed so far were either tested on a single dataset, or on datasets containing a similar type of X-ray imagery. Furthermore, there are limited frameworks which are applied to the complex radiographs for the detection of heavily occluded and cluttered baggage items. Many latest frameworks that can detect multiple objects and potential anomalies use CNN models (and object detectors) as a black-box, where the raw images are passed to the network for object detection. Considering the real-world scenarios where most of the baggage items are heavily occluded that even human experts can miss them, it will not be straightforward for these frameworks to produce optimal results. As reported in \cite{_46}, for a deep network to estimate the correct class, the feature kernels should be distinct. This condition is hard to fulfil for occluded objects obtained through raw images (without any initial processing), making thus the prediction of the correct class quite a challenge \cite{_46}. In \cite{_32}, different CNN based object detectors were evaluated on the SIXray10 subset to only detect \textit{guns} and \textit{knives}. 
Despite the progress accomplished by the above works, the challenge of correctly recognizing heavily occluded and cluttered items in SIXray dataset scans is still to be addressed.

\section{Contribution}
\noindent  In this paper, we present a cascaded structure tensor (CST) framework for the detection and classification of suspicious items. The proposed framework is unique as it only uses a single feed-forward CNN model for object recognition, and instead of passing raw images or removing unwanted regions, it intelligently extracts each object proposal by iteratively picking the contour-based transitional information from different orientations within the input scan. The proposed framework is robust to occlusion and can easily detect heavily cluttered objects as it will be evidenced in the results section. The main contributions of this paper and the features of the proposed framework are summarized below: 
\begin{itemize}[leftmargin=*]
  \item    We present a novel object recognition framework that can extract suspicious baggage items from X-ray scans and recognizes them using just a single feed-forward CNN model.
\item    The proposed framework is immunized to the class imbalance problem since it is trained directly on the balanced set of normal and suspicious items proposals rather than on the set of scans containing an imbalanced ratio of normal and suspicious items.
\item    The extraction of items proposals is performed through a novel CST framework, which analyzes the object transitions and coherency within a series of tensors generated from the candidate X-ray scan. 
\item    The proposed CST framework exhibits high robustness to occlusion, scan type, noisy artefacts and to highly cluttered scenarios.
\item    The proposed framework achieves mean intersection-over-union ($IoU$) score of 0.9644 and 0.9689, area under the curve ($AUC$) score of 0.9878 and 0.9950, and a mean average precision ($\mu_{AP}$) score of 0.9343 and 0.9595 for detecting normal and suspicious items from GDXray and SIXray dataset, respectively (see Section \ref{sec:resuts}). 
\item The proposed framework achieves 4.76\% better run-time performance, compared to existing state-of-the-art solutions such as \cite{_29}, \cite{_30}, \cite{_31}, \cite{_32}, \cite{_42} and \cite{_44} which are based on exhaustive searches and anchor box estimations. 

\end{itemize}

\noindent The rest of the paper is organized as follows: Section \ref{sec:prop} describes the proposed framework and its implementation, Section \ref{sec:expsetuo} presents the experimental setup, Section \ref{sec:resuts} reports the  experiments, the   results, and  a comparison with state-of-the-art solutions. Finally, Section \ref{sec:conc} concludes the paper.

\section{Proposed Method} \label{sec:prop}
\noindent The block diagram of the proposed framework is depicted in Figure \ref{fig:block2}.  In the first step, we enhance the contrast of the input image via an adaptive histogram equalization \cite{_47}. Afterwards, we generate a series of tensors where each tensor embodies information about the targeted objects from different orientations. We employ these tensors for the automatic extraction of baggage items proposals from the scans. We subsequently pass these proposals to a  pre-trained network for object recognition. Each module within the proposed framework is described next:

\subsection{Preprocessing}
\noindent 
The primary objective of the preprocessing stage is to enhance the low contrasted input. We perform contrast stretching through adaptive histogram equalization \cite{_47}.  Let $\xi_{X} \in \Re^{M \times N}$  the X-ray scan where $M$ and $N$ denotes the number of rows and columns, respectively. Let $\wp$  be an arbitrary patch of $\xi_{X}$  where $\wp$ is obtained by dividing $\xi_{X}$ into $I \times J$ grid of rectangular patches. The histogram of $\wp$ is computed and is locally normalized using the following relation:
\begin{equation}
\hbar=\Gamma\left(\frac{\Delta_{\wp}-\Delta_{\wp min}}{M*N-\Delta_{\wp min}}(L_{M}-1)\right)
\end{equation}
\noindent where $\Delta_{\wp}$ is the cumulative distribution function of $\wp$, $\Delta_{\wp min}$ is the minimum value of $\Delta_{\wp}$, $L_{M}$ represents the maximum grayscale level of $\xi_{X}$, $\Gamma(.)$ is the rounding function and  $\hbar$ is enhanced histogram for $\wp$. This process is repeated for all the patches of $\xi_{X}$ to obtain the contrast stretched version $\xi_{X}^t$  as shown in Figure 3. 
We can observe that the occluded \textit{gun} is visible in the enhanced scan.
\begin{figure}
\includegraphics[scale=0.22]{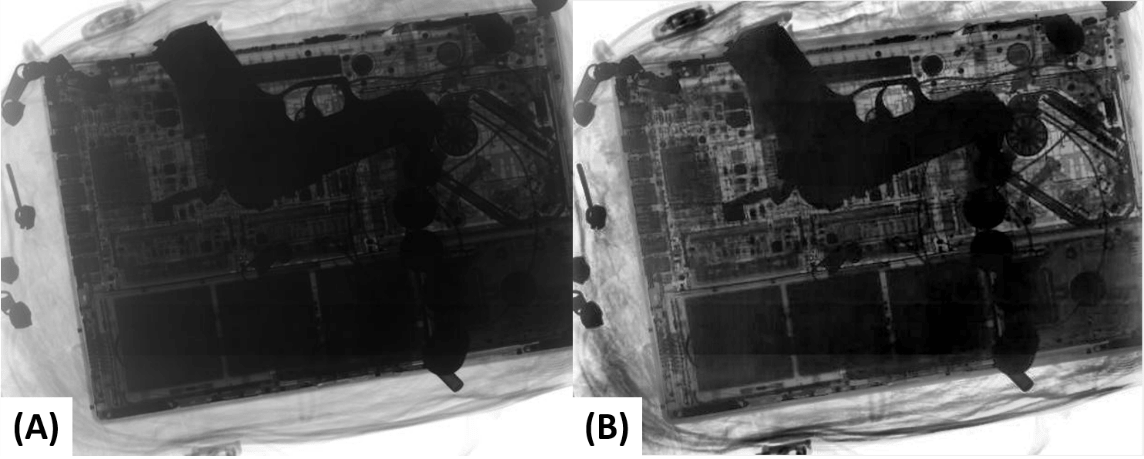}
\caption{\small Preprocessing stage: (A) original image $\xi_{X}$ containing an occluded \textit{gun}, (B) enhanced image $\xi_{X}^t$ in which the \textit{gun} is visible. }
\centering
\label{fig:block3}
\end{figure}
\subsection{Cascaded Structure Tensor (CST) Framework}
\noindent  CST is a framework for object extraction from a candidate scan which is based on a  novel contoured structure tensor approach. We will first report the conventional structure tensor; then describe our more generalized approach proposed in this context of suspicious baggage items detection.
\subsubsection{Conventional 2D Discrete Structure Tensor}
\noindent A 2D discrete structure tensor is defined as a second moment matrix derived from the image gradients. It reflects the predominant orientations of the changes (contained in the gradients) within a specified neighbourhood of any pixel in the input scan \cite{_69,_70}.  Furthermore, structure tensor can  tell the degree to which those orientations are coherent. For a pixel  $p$ in $\xi_{X}^t$, the structure tensor $S(p)$ is defined as:
\begin{equation}
\resizebox{0.9\hsize}{!}{
$S(p)=$
$\begin{bmatrix}
  \sum_{r}\varphi_{(r)}(\nabla^{x}_{(p-r)})^2 & \sum_{r}\varphi_{(r)}(\nabla^{x}_{(p-r)}.\nabla^{y}_{(p-r)}) \\
  \sum_{r}\varphi_{(r)}(\nabla^{y}_{(p-r)}.\nabla^{x}_{(p-r)}) &
  \sum_{r}\varphi_{(r)}(\nabla^{y}_{(p-r)})^2
\end{bmatrix}$}
\end{equation}
\noindent where $\varphi$ is a smoothing filter typically chosen as a Gaussian function, $\nabla^{x}$ and $\nabla^{y}$ denotes the image gradients w.r.t $x$  and $y$ direction, respectively. From Eq. (2), we can observe that for each pixel in the input scan, we obtain a $2\times2$ matrix representing the distribution of orientations within the image gradients w.r.t its neighbourhood. In order to measure how strongly this matrix is biased towards a particular direction, the degree of coherency or anisotropy is computed through the two eigenvalues from $S(p)$  as shown below:

\begin{equation}
c=\left(\frac{\lambda_1-\lambda_2}{\lambda_1+\lambda_2}\right)^2
\end{equation}
\noindent where $c$  represents the coherence measure such that $0\le c \le 1$, $\lambda_1$ and $\lambda_2$ are the two eigenvalues. In the proposed framework, we are primarily not interested in finding out the strong orientations within the specified neighborhood of any pixel. Here, our purpose is to extract the coherence maps such that they represent the maximum transitions of baggage items from different orientations. Furthermore, in the conventional structure tensor, the gradients are computed in two orthogonal directions only. However, the baggage items within X-ray scans can have other directions as well.
\begin{figure}[htb]
\begin{center}
\includegraphics[scale=0.255]{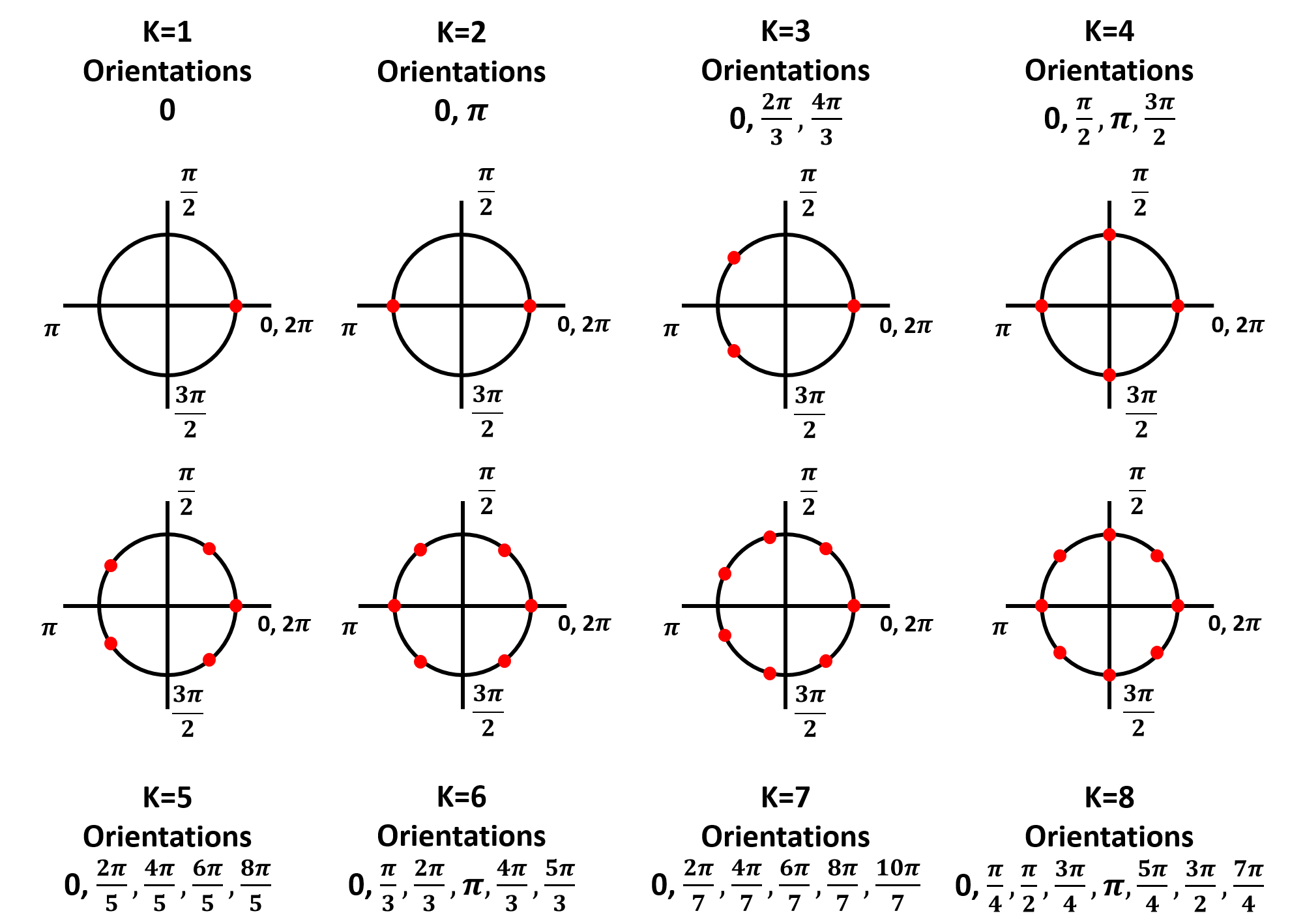}
\end{center}
\caption{\small Gradient orientations for computing tensors in CST framework. }
\centering
\label{fig:block4}
\end{figure}
\subsubsection{Modified Structure Tensor}
\noindent{The modified version of the structure tensor can reveal the objects oriented in any direction within the candidate image. It is defined over $K$ image gradients associated to $K$ number of different orientations yielding the structure tensor matrix of $K\times K$ order where $K \in \mathbb{N}$ as shown below:}

\begin{equation}
\Im =
\begin{bmatrix}
  \Im_0^0 & \Im_1^0 & \cdots & \Im_{K-1}^0 \\
  \Im_0^1 & \Im_1^1  & \cdots & \Im_{K-1}^1 \\
  \vdots & \vdots & \ddots & \vdots \\
  \Im_0^{K-1} & \Im_1^{K-1} & \cdots & \Im_{K-1}^{K-1} \\
\end{bmatrix}
\end{equation}

\begin{figure}
\includegraphics[scale=0.0825]{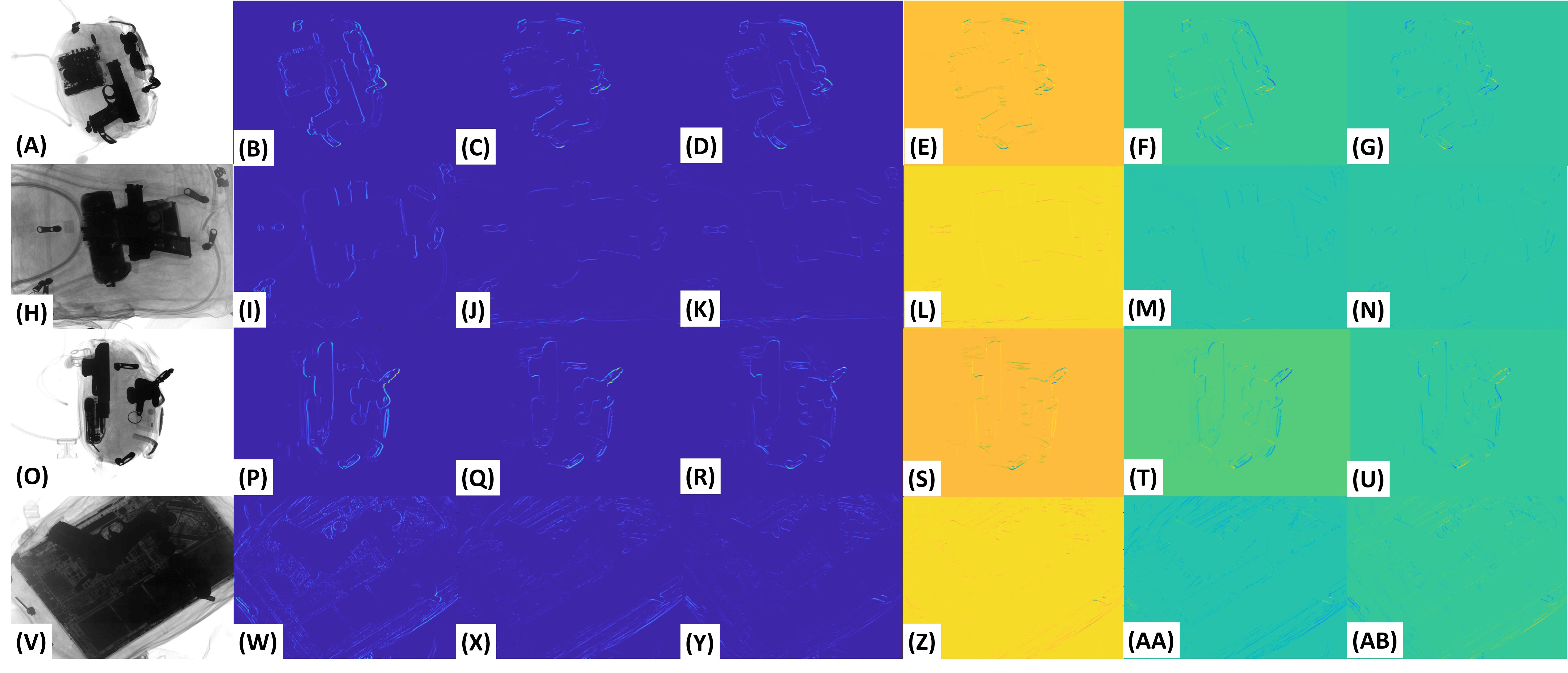}
\caption{\small Six unique tensors from $\Im$ for $K=3$. For each image (in the first column), the respective tensors are arranged from left to right according to their norm. }
\centering
\label{fig:block5}
\end{figure}

\noindent{Each coherence map or tensor ($\Im_j^i$) in Eq. (4) is an outer product of smoothing filter and the image gradients i.e. $\Im_j^i=\varphi * (\nabla^i . \nabla^j)$  having  $M \times N$ dimension, $\nabla^i$ and $\nabla^j$ are the image gradients in the direction $i$ and $j$, respectively. Also, rather than using a Gaussian filter for smoothing, we employ anisotropic diffusion filter \cite{_52} as it is extremely robust in removing noisy outliers while preserving the edge information. Moreover, it can be noted that $\Im$ is a symmetric matrix which means that $\Im_j^i=\Im_i^j$ or in other words there are $\frac{K(K+1)}{2}$  unique tensors in $\Im$. The gradient orientations ($\vartheta$) in the modified structure tensor is computed through $\vartheta = \frac{2 \pi k}{K}$ where $k$ is varied from $0$ to $K-1$. For example, for $K=1$, we have only one tensor for one image gradient oriented at $0$ rad. For $K=2$, we have four tensors for two image gradients oriented at $0$ rad and $\pi$ rad.  Figure \ref{fig:block4} shows the orientation of image gradients for different values of $K$.  Figure \ref{fig:block5} shows the six unique tensors obtained for randomly selected scans when $K=3$.}

\subsubsection{Coherent Tensors}
\noindent As mentioned above, with $K$ orientations,  we obtain $\frac{K(K+1)}{2}$ unique tensors. Each tensor highlights objects in the candidate image w.r.t the predominant orientations of its respective image gradients. However, considering all these tensors for the generation of object contours (and proposals) is redundant, time-consuming, and might make the proposed framework vulnerable to noise. We instead select a reduced number of $M$   predominant tensors, which we dubbed the Coherent Tensors. The optimal values of $K$ and $M$ will be determined empirically in the ablation analysis (see Section \ref{ablationStudy}).
\subsubsection{Extraction of Object Proposals}
\begin{figure}
\includegraphics[scale=0.102]{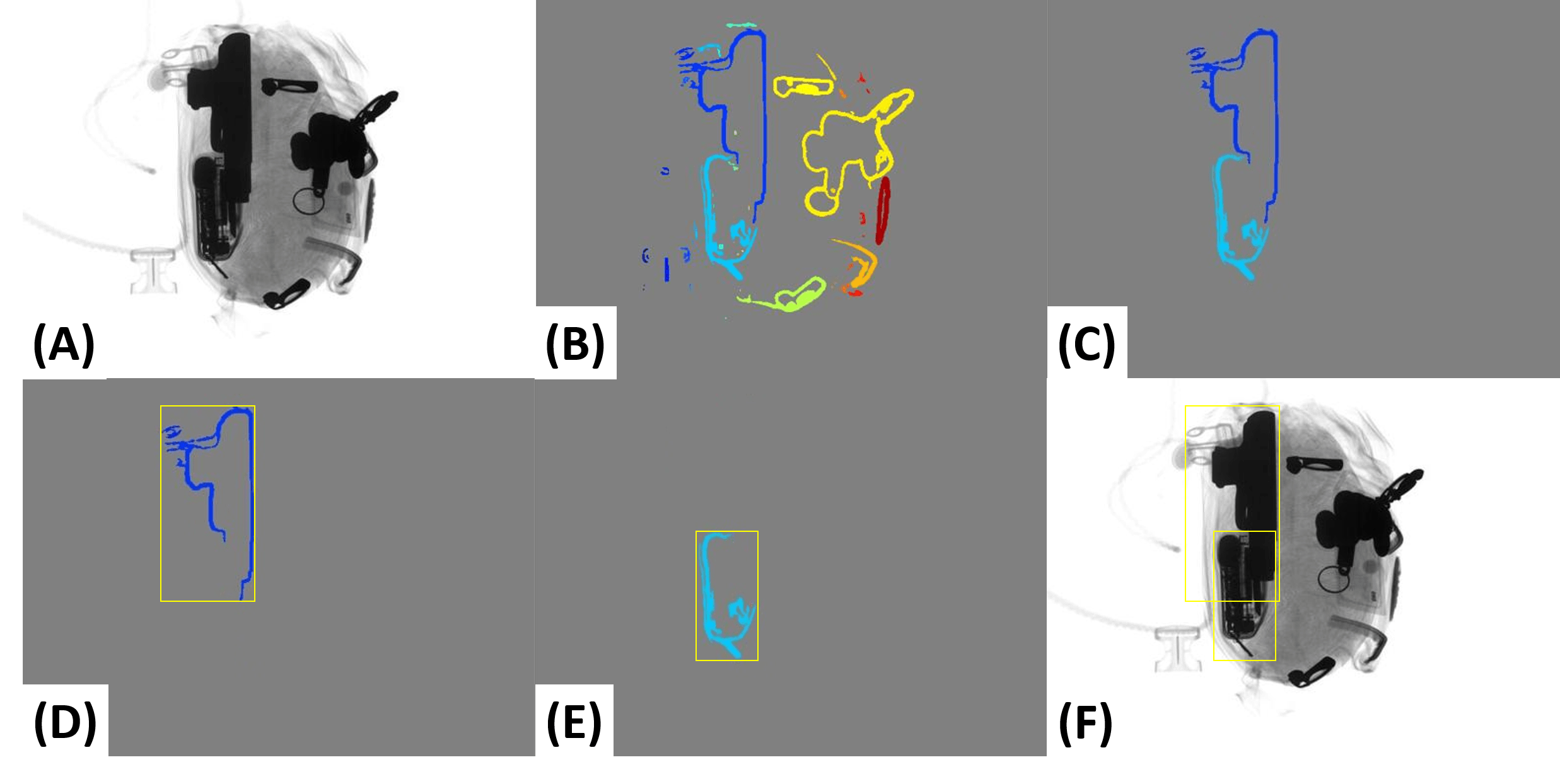}
\caption{\small Bounding box generation for each occluded object: (A) original image, (B) CST contour map, (C) contours of \textit{pistol} and the \textit{chip} objects. (D) bounding box generated for \textit{pistol} object, (E) bounding box generated for the \textit{chip} object, (F) bounding boxes overlaid on the original scan for the extracted objects. }
\centering
\label{fig:bounBoxExtract}
\end{figure}

\begin{figure}
\begin{center}
\includegraphics[scale=0.239]{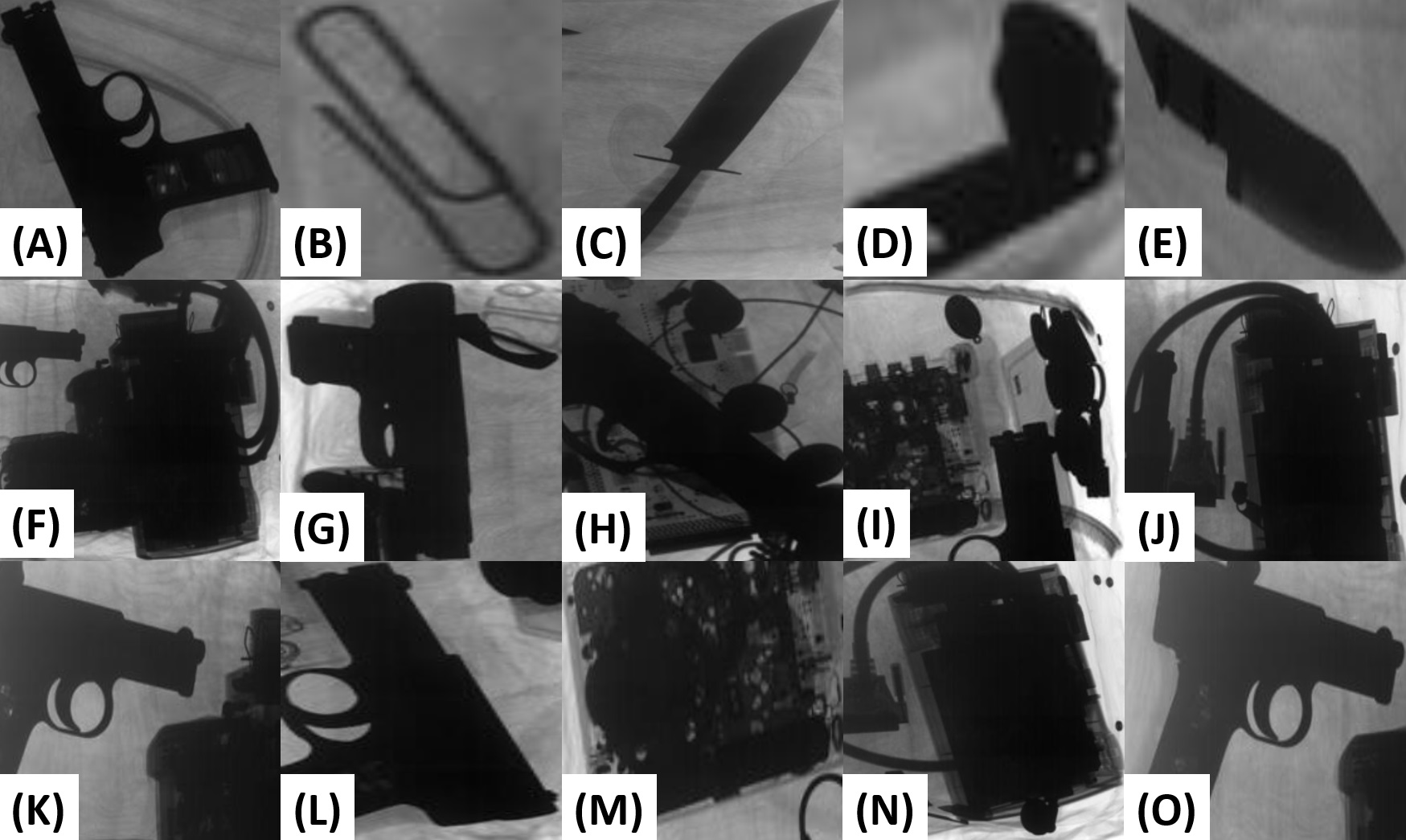}
\end{center}
\caption{\small Examples of extracted proposals. The top row shows proposals for isolated objects and bottom two rows show proposals of the occluded, merged and cluttered objects. }
\centering
\label{fig:block8}
\end{figure}

\noindent 
After obtaining the coherent tensors, we add them together to obtain a single coherent representation ($\Im_{\Theta}$) of the candidate image. Then, $\Im_{\Theta}$  is binarized and morphologically enhanced to remove the unwanted blobs and noisy artefacts. 
Moreover, after generating the baggage items contours, they are labelled through connected component analysis \cite{_53} and for each labelled contour, a bounding box is generated based upon minimum bounding rectangle technique (see  Figure \ref{fig:bounBoxExtract}). This bounding box is then used in the extraction of the respective object proposal from the candidate image. 
Proposals generated by the CST framework contain either a single object
or overlapping objects (see examples in Figure \ref{fig:block8}). 

\begin{figure}
\begin{center}
\includegraphics[scale=0.155]{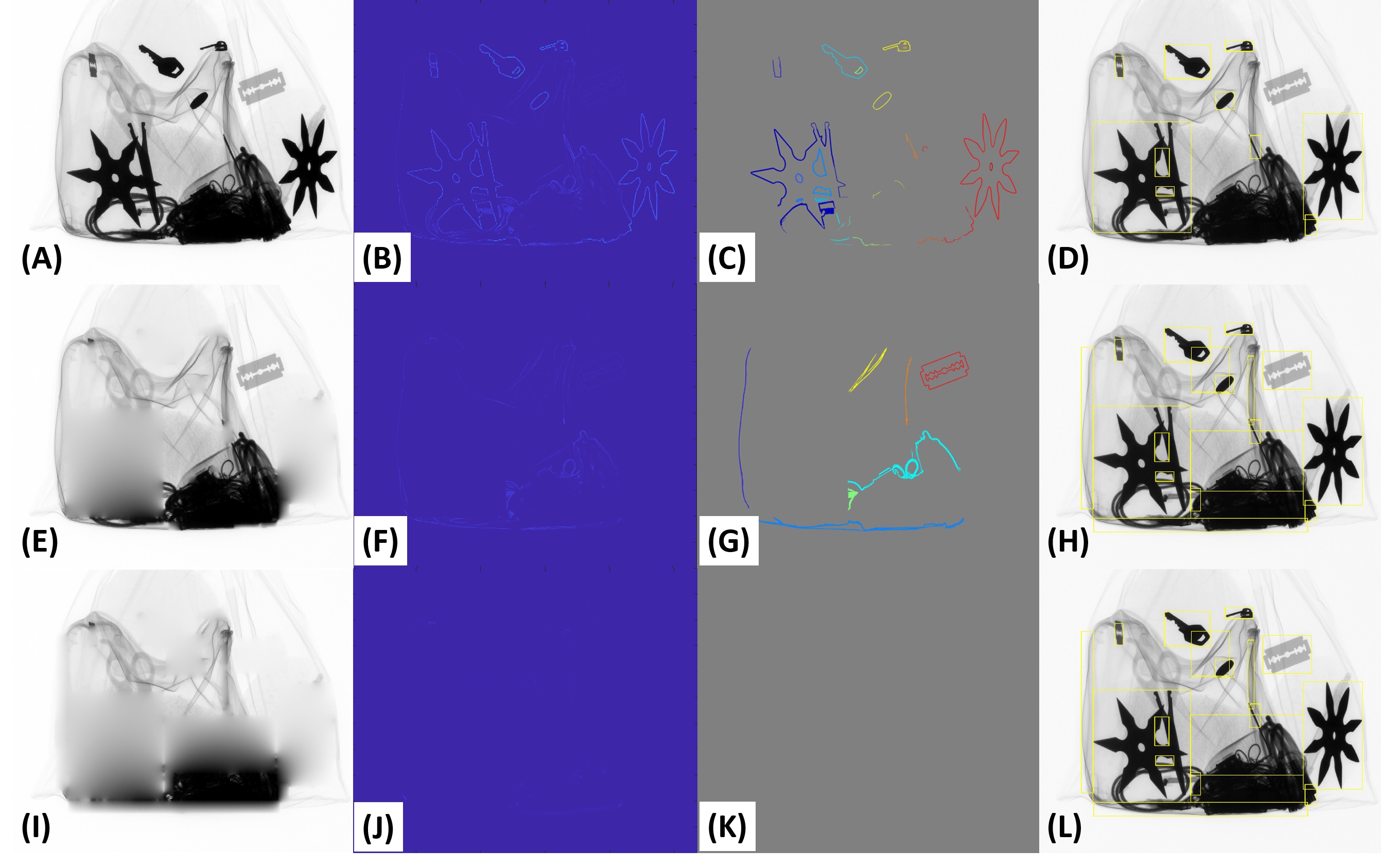}
\end{center}
\caption{\small CST framework convergence. First row shows the first iteration, second row shows the second iteration whereas third row shows the third and final iteration for extracting the objects from the candidate scan. }
\centering
\label{fig:CSTmultipass}
\end{figure}

\noindent 
Object borders in an X-ray scan exhibit different level of intensity. This disparity make objects with high transition-level borders (e.g. \textit{shuriken} in Figure \ref{fig:CSTmultipass}A)  have more detection likelihood  compared to those showing weaker transitions (\textit{razor blades} in Figure \ref{fig:CSTmultipass}A).
In order to accommodate these different levels of boundary intensity, we employ the CST in a multi-pass scheme in which regions corresponding to the proposals detected in the previous pass 
(most likely having a high boundary transition level) are  discarded from the $\xi_{X}^t$ image, 
by computing the discrete Laplacian and then solving the Dirichlet boundary value problem \cite{_68}, in the subsequent pass. In this way, the pixels in the proposals bounding boxes of the previous pass are  replaced with the intensity values derived from the proposal boundaries (see Figure   \ref{fig:CSTmultipass} E and I). This iterative process is repeated until there are no more object transitions left to be picked from the candidate scan as shown in Figure \ref{fig:CSTmultipass} (K).  The detailed pseudo-code of the proposed CST framework is presented below:
\begin{algorithm}
\SetAlgoLined
\DontPrintSemicolon
\textbf{Input: }Enhanced Image $\xi_{X}^t$, No. of Orientations $K$

\textbf{Output: }Object Proposals $O^\rho_\Upsilon$

\textit{hasObjects} $\gets$ true

$O^\rho_\Upsilon$ $\gets$ $\phi$

\While{\textit{hasObjects} is true}{

Compute $K^2$ tensors from $\xi_{X}^t$

Generate the coherent tensor representation $\Im_{\Theta}$

Use $\Im_{\Theta}$ to compute contour map $\Phi_{\Theta}$

Compute a labeled object map $\ell_\Phi$ from $\Phi_{\Theta}$

\If{$\ell_\Phi$ is $\phi$}{\textit{hasObjects} $\gets$ false}

\ForEach {label $\ell$ in $\ell_\Phi$}{

   Compute its bounding box $\beta_B$
   
   Use $\beta_B$ to crop proposal $O^\rho$ from $\xi_{X}^t$
   
   Add $O^\rho$ in the list of proposals $O^\rho_\Upsilon$
   
   Use $\beta_B$ to in-paint $O^\rho$ in $\xi_{X}^t$
}
}
 \caption{Proposed CST Framework}
\end{algorithm}

\subsubsection{Object Recognition}
\noindent{After extracting the object proposals, these are passed to the pre-trained ResNet\textsubscript{50} model \cite{_48}  for recognition. The ResNet\textsubscript{50} exhibits good performance in catering the vanishing gradient problem through the residual blocks \cite{_49}.
We employ ResNet\textsubscript{50} in fine-tuning mode, whereby we replace the final classification layer with our custom layer for the recognition of proposals within this application. We do not freeze the rest of the layers so that they also get updated during the training phase to recognize the object proposals effectively. However, we use the original weights in the initialization phase for faster convergence. 
The training set is composed of object proposals obtained with the CST framework, which generates around 167 proposals on average per scan. This amplification in the number of training samples allows deriving balanced sets for normal and suspicious items, as will be described further in the experimental setup in Section \ref{sec:expsetuo}. }
In the training set, proposals are labelled as follows:
1) Proposals containing a single suspicious item or a portion of a single item are labelled by that item category  
2) Proposals containing overlapped suspicious items are labelled with the item category  occupying the largest area in the bounding box.
3) Proposals that do not contain suspicious items are labelled as normal.

\section{Experimental Setup} \label{sec:expsetuo}
\noindent The proposed framework is evaluated against state-of-the-art methods on different publicly available datasets using a variety of evaluation metrics. In this section, we have listed the detailed description of all the datasets and the evaluation metrics. This section also describes the training details of the pre-trained CNN model.

\subsection{Datasets}
\noindent We assessed the  proposed system  with  GDXray \cite{_55} and SIXray \cite{_46} datasets, each of which is explained below:
\subsubsection{GRIMA X-ray Database}
The GRIMA X-ray Database (GDXray) \cite{_55} was acquired with an X-ray detector (Canon CXDI-50G) and an X-ray emitter tube (Poksom PXM-20BT). It contains 19,407 X-ray scans arranged in welds, casting, baggage, nature and settings categories. In this paper, we only use the baggage scans to test the proposed framework, as this is the only relevant category for suspicious items detection. The baggage group has 8,150 X-ray scans containing both occluded and non-occluded items. Apart from this, it contains the marked ground truths for \textit{handguns}, \textit{razor blades}, and \textit{shuriken}.  For a more in-depth evaluation, we have refined this categorization by splitting the original \textit{handgun} category into two classes; namely, \textit{pistol} and \textit{revolver}.  We have also identified and annotated three new classes i.e. the \textit{knife}, \textit{mobile phone} class and the \textit{chip} class, which represents all the electronic gadgets, including \textit{laptops}. We adopted a training set in accordance with standard defined in \cite{_55}, i.e. 400 scans from B0049, B0050 and B0051 series containing proposals for \textit{revolver} (\textit{handgun}), \textit{shuriken} and \textit{razor blades}, to which we added 388 more scans for the new categories (\textit{chip}, \textit{pistol}, \textit{mobile} and \textit{knives}).  From the 788 training scans, we have obtained a total number of 140,264 proposals which were divided into suspicious and normal categories. The last, is an auxiliary class that includes proposals of miscellaneous and unimportant items like \textit{keys} and \textit{bag zippers}, which are generated by CST framework. However, we discarded 84,162 normal proposals to keep the number of suspicious and unsuspicious items balanced. The detailed summary of the GDXray dataset is depicted in Table I.
\begin{table}[htb]
    \centering
    \caption{Detailed description of GDXray  dataset.}
    \begin{tabular}{c c c c}
        \toprule
         \textbf{Total Scans} &  \textbf{Dataset Split} & \textbf{Training Proposals} &  \textbf{Items}\\
        \hline
            8,150 & Training: 788\textsuperscript{\#} & Total: 140,264 & Pistol** \\
            & Testing: 7,362 & Normal: 28,049 & Revolver** \\
            & & Suspicious: 28,053 & Shurken \\ 
            & & Discarded: 84,162\textsuperscript{$\star$} & Knife* \\
            & & \textbf{Average: 178 / scan} & Razor Blades \\
            & & & Chip* \\
            & & & Mobile* \\
         \toprule
    \end{tabular}
    \begin{tablenotes}[flushleft]
        \item[*] * These items have been identified locally for  more in-depth validation of the proposed framework. \textit{Chip} class represents all the electronic gadgets including \textit{laptops} (except \textit{mobiles}).
        \item[$\star$] $\star$ these are the normal proposals that have been discarded for balanced training of the CNN model for proposals recognition.
        \item[**] ** The original handgun category is further broken down into \textit{pistol} and \textit{revolver} because both items are found to be in abundance within this dataset.
        \item[\#] \# 400 scans from B0049, B0050 and B0051 series are used for extracting \textit{revolver}, \textit{shuriken} and \textit{razor blades} as per the criteria defined in \cite{_55} and using this same percentage of training and testing split, we used 388 more scans to train the model for the extraction of other items.
    \end{tablenotes}
    \label{tab:my_label3}
\end{table}

\subsubsection{Security Inspection X-ray Dataset}
\noindent 
Security Inspection X-ray (SIXray) \cite{_46} is one of the largest datasets for the detection of heavily occluded and cluttered suspicious items. It contains 1,059,231 color X-ray scans having 8,929 suspicious items which are classified into six groups i.e. \textit{gun}, \textit{knife}, \textit{wrench}, \textit{pliers}, \textit{scissor} and \textit{hammer}. All the images are stored in JPEG format, and the detailed description of the dataset is presented in Table II. To validate the performance of the proposed framework against the class imbalance problem, the same subsets have been utilized, as described in \cite{_46}, in which the ratio of suspicious items and normal objects have been matched with real-world scenarios. Also, the ratio of 4 to 1 for training and testing has been maintained in accordance with \cite{_46}. 
For the SIXray dataset, we also have added a separate \textit{normal} class to filter the proposals of miscellaneous and unimportant items that are generated by the CST framework. Moreover, it should also be noted from Table I and II that for each dataset we have trained the classification model on the balanced set of normal and suspicious items proposals where the excessive normal items proposals are discarded to avoid the classifier bias.
\begin{table}[htb]
    \centering
    \caption{Detailed description of SIXray dataset}
    \begin{tabular}{c c c c}
        \toprule
         \textbf{Subsets\textsuperscript{*}} & \textbf{Scans}\textsuperscript{$\dagger$} &  \textbf{Dataset Split} & \textbf{Training Proposals} \\
        \hline
            SIXray10 & 98,219 & Training: 78,575 & Total: 12,179,125 \\
            & & Testing: 19,644  & Normal: 2,435,819 \\
            & & & Suspicious: 2,435,825 \\ 
            & & & Discarded: 7,307,481\textsuperscript{$\star$} \\
            \hline
            SIXray100 & 901,829 & Training: 721,463 & Total: 111,826,765 \\
            & & Testing: 180,366   & Normal: 22,365,348 \\
            & & & Suspicious: 22,365,353 \\ 
            & & & Discarded: 67,096,064\textsuperscript{$\star$} \\
            \hline
            SIXray1k & 1,051,302 & Training: 841,042 & Total: 130,361,510 \\
            & & Testing: 210,260   & Normal: 26,072,296 \\
            & & & Suspicious: 26,072,302 \\ 
            & & & Discarded: 78,216,912\textsuperscript{$\star$} \\
            & & & \textbf{Average: 155 / scan} \\
         \toprule
    \end{tabular}
    \begin{tablenotes}[flushleft]
        \item[*] * The suspicious items within the SIXray dataset are: \textit{gun}, \textit{knife}, \textit{wrench}, \textit{pliers}, \textit{scissor} and \textit{hammer}.
        \item[$\star$] $\star$ These are the normal proposals that have been discarded for balanced training of the CNN model for proposals recognition.
        \item[$\dagger$] $\dagger$ total scans within the SIXray dataset are 1,059,231 in which 8,929 scans are positive (containing one or more suspicious items), and the rest of 1,050,302 scans are negative.
    \end{tablenotes}
    \label{tab:my_label3}
\end{table}
\subsection{Training Details}
\noindent{The classification of the baggage items is performed through pre-trained ResNet\textsubscript{50} model after fine-tuning it on the object proposals extracted from the scans of GDXray and SIXray datasets.}

\begin{figure}
 \begin{center}
 \includegraphics[scale=0.133]{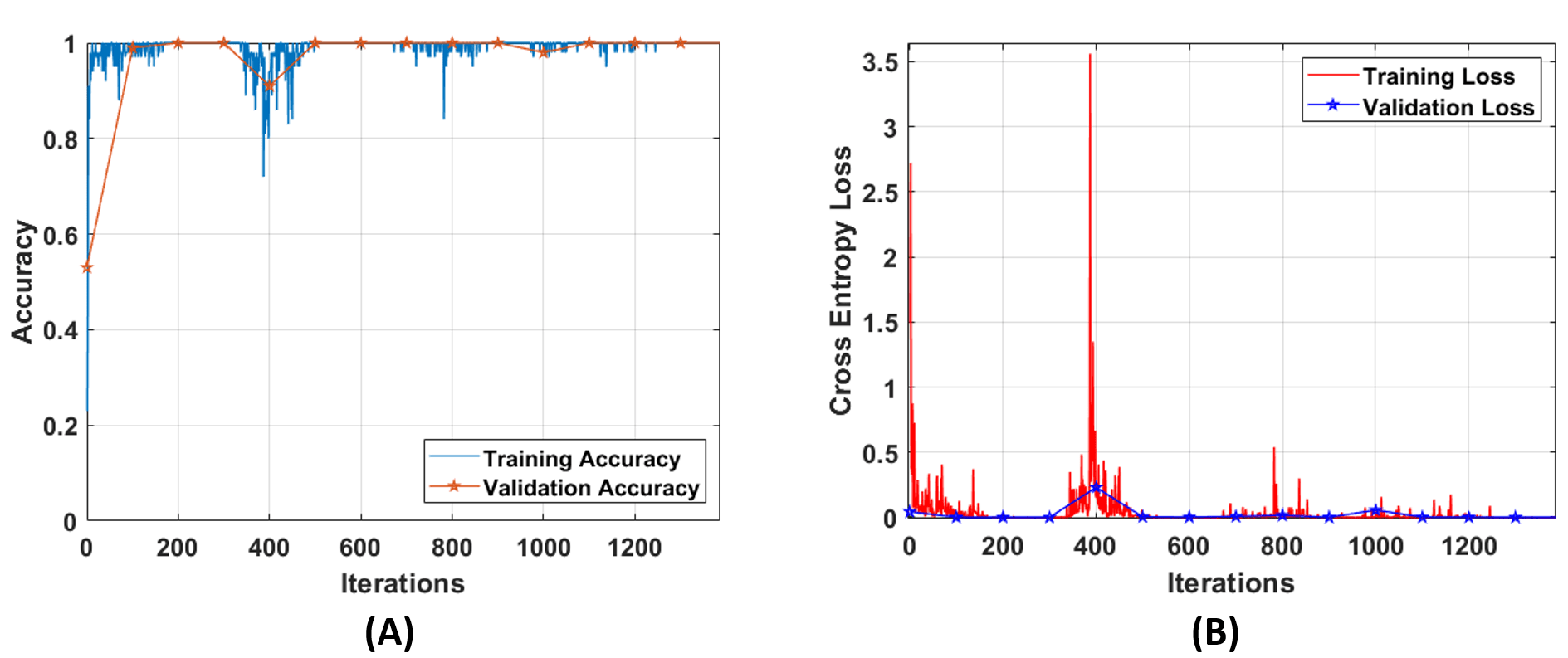}
 \end{center}
 \caption{\small Training performance of the proposed framework in terms of  accuracy (A) and  cross entropy loss (B). First 1300 iterations, after which the convergence was achieved,  are shown.
 }
 \centering
 \label{fig:block10}
 \end{figure}

\noindent The training process was conducted for 13 epochs having a mini-batch size of 64 using MATLAB R2019a with deep learning toolbox, on a machine with an Intel Core i5-8400@2.8GHz processor, 16 GB RAM and NVIDIA RTX 2080 GPU with cuDNN 7.5. The optimization during the training phase was performed through ADAM \cite{_56} with a base learning rate of 0.001, whereas 20\% of the training data is used for validation which is performed after 100 consecutive iterations. Furthermore, the categorical cross-entropy loss function ($\zeta_L$)  is employed during training which is computed through Eq. (5):
\begin{equation}
\zeta_L = - \sum_{i=1}^{s} \sum_{j=1}^{\omega} y_{i,j} log(p_{i,j})
\end{equation}
where $s$ denotes the total number of samples, $\omega$ denotes total number of classes, $y_{i,j}$ is a binary indicator stating whether $i^{th}$ sample belongs to $j^{th}$ class and $p_{i,j}$ is the predicted probability of the $i^{th}$ sample for $j^{th}$ class. Figure \ref{fig:block10} shows the training performance of the proposed framework.
\subsection{Evaluation Criteria} \label{sec:criteria}
\noindent{The performance is evaluated based on the following metrics:
\subsubsection{Intersection over Union}
$IoU$ describes the overlapping area between the extracted object bounding box and the corresponding ground truth. It is also known as the Jaccard’s similarity index and it is computed through Eq. (6):}

\begin{equation}
    IoU = \frac{\Lambda (\beta_B \cap G_\tau)}{\Lambda (\beta_B \cup G_\tau)}
\end{equation}

\noindent where $\beta_B$ is the extracted bounding box for the baggage item, $G_\tau$ is the ground truth and $\Lambda (.)$ computes the area of the passed region. Although $IoU$ measures the ability of the proposed CST framework to extract the suspicious items, it does not measure the detection capabilities of the proposed framework.

\subsubsection{Accuracy}
\noindent Accuracy describes the performance  of correctly identify an object and no-object regions as described 
\begin{equation}
    Accuracy = \frac{T_P + T_N}{T_P + F_N + T_N + F_P}
\end{equation}
where $T_P$ denotes the true positive samples, $T_N$ denotes the true negative samples, $F_P$ denotes the false positive samples and $F_N$ denotes the false negative samples. 

\subsubsection{Recall}
Recall, or sensitivity is the true positive rate ($T_{PR}$) which indicates the completeness of the proposed framework to correctly classifying the object regions. 
\begin{equation}
    T_{PR} = \frac{T_P}{T_P + F_N}
\end{equation}
\subsubsection{Precision}
Precision ($P_R$) describes the purity of the proposed framework in correctly identifying the object regions against the ground truth.
\begin{equation}
    P_{R} = \frac{T_P}{T_P + F_P}
\end{equation}
\subsubsection{Average Precision}
Average precision ($\lambda_P$) represents the area under the precision-recall ($P_{RC}$) curve. It is a measure that indicates the ability of the proposed framework to correctly identifying positive samples (object proposals in our case) of each class or group. $\lambda_P$ (for each class) is computed by sorting the images based on confidence scores and marking the bounding box predictions positive or negative. The prediction is marked positive if $IoU \geq 0.5$ and negative otherwise.  Afterwards, the $T_{PR}$ and $P_R$, computed using Eq. (8) and (9), are used to generate $\lambda_P$ as follows:
\begin{equation}
    \lambda_P = \int_{a}^{b}\hat{P}_R(r)dr \approx \left( \sum_{r = a}^{b} P_R^I(r) \nabla T_{PR}(r) \right)
\end{equation}
\begin{equation}
    P_R^I(r) = \underset{\hat{r} \geq r}{max}\,  P_R(\hat{r}) 
\end{equation}
\noindent where $P_R^I$ denotes the estimated precision, $\nabla T_{PR}$ represents the change (difference) in the consecutive recall values, $a$ is the starting point which is equal to 0, $b$ is the endpoint of the interpolation interval which is 1 and $\hat{P}_R$ represents the actual precision curve. After computing the $\lambda_P$ for each class, the $\mu_{AP}$ score is computed using Eq. (12):
\begin{equation}
   \mu_{AP}=\frac{1}{N_c} \left( \sum_{k=0}^{N_c - 1} \lambda_P(k) \right)
\end{equation}

\noindent where $N_c$ denotes the number of classes in each dataset.

\subsubsection{$F_1$ Score}
$F_1$ score measures the capacity of   correctly classifying  samples in a highly imbalanced scenario. It is computed by considering both  $P_R$ and $T_{PR}$ scores. 
\begin{equation}
   F_1 = 2 \times \frac{P_R \times T_{PR}}{P_R + T_{PR}}
\end{equation}

\subsubsection{Receiver Operator Chracteristics (ROC) Curve}
ROC curves indicate the degree of how much the proposed framework confuses between different classes.  It is computed with the true positive rate $T_{PR}$ and the false positive rate ($F_{PR}$).
\begin{equation}
   F_{PR} = 1 - Specificity = 1 - \frac{T_N}{T_N + F_P} = \frac{F_P}{T_N+F_P}
\end{equation}
where specificity is the true negative rate.  After computing the ROC curves for each class within GDXray and SIXray datasets, $AUC$ is computed by numerically integrating the ROC curve.
\subsubsection{Qualitative Evaluations}
\noindent Apart from the quantitative evaluations, the proposed framework is thoroughly validated through qualitative evaluations as will be shown in the results section.
\section{Results} \label{sec:resuts}
\noindent In this section, we report the results obtained through a comprehensive series of experiments conducted with GDXray and SIXray datasets. In Section \ref{ablationStudy}, we report the ablative analysis we conducted to  study the effects of the number of orientations, the number of  coherent tensors, and  the choice of classification network on the overall system performance.  
In Section \ref{system-validation}, we present the performance of the proposed system evaluated based on the metrics explained in Section \ref{sec:criteria}. Then, in Section \ref{comparative-study}, we report a comparative study with state-of-the-art methods. In Section \ref{failure-cases}, we discuss  the limitations of the proposed framework.

\subsection{Ablation Study} \label{ablationStudy}
\noindent The number of orientation  $K$ and the number of selected coherent tensors  $M$ are the main hyper-parameters of the \textit{CST} framework.  The number $K$ is related to the capacity of the CST for accommodating objects oriented in different directions, while $M$ is related to the most relevant tensors. 

\begin{figure}
\begin{center}
\includegraphics[scale=0.09]{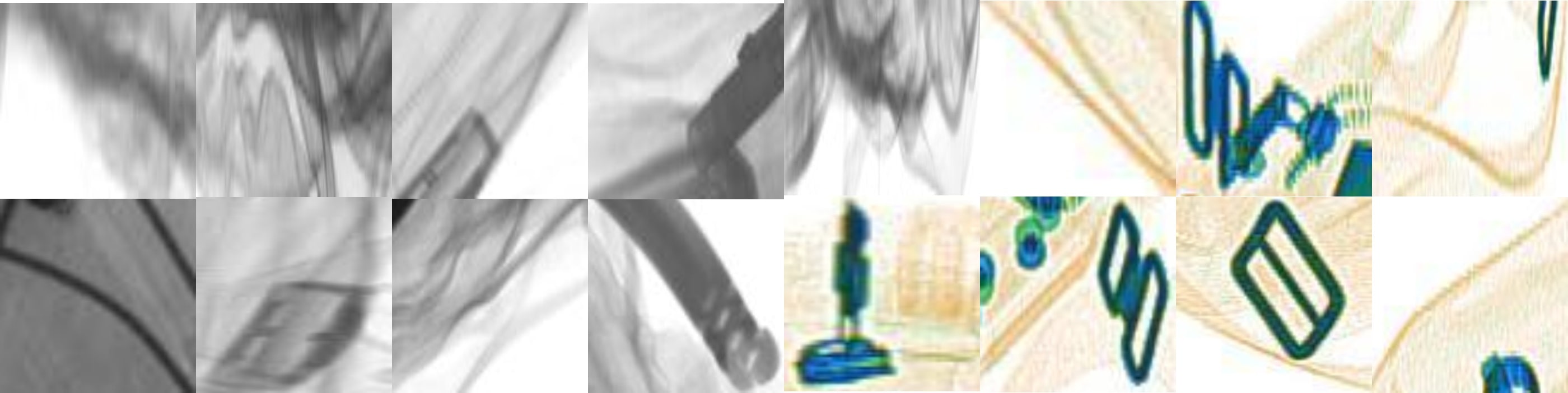}
\end{center}
\caption{\small Noisy and miscellaneous proposals. }
\centering
\label{fig:block17}
\end{figure}

\noindent In the first experiment, we varied the number of orientations  $K$ from 2 to 4, and we computed  $\mu_{AP}$  for a number of selected tensors  $M$ varying from 1 to  $\frac{K(K+1)}{2}$.  In the results, reported in
Table \ref{tab:ablationKM}, we notice an overall enhancement of the performance as $K$ increases.  We also notice that for $K=4$, the performance reaches a  peak of 0.934  and  0.959, at  $ M=2$, for both GDXray and  SIXray,  then starts to decrease. This decay can be explained by the fact that including more tensors adds spikes and noisy transitions leading to the generation of noisy edges and negative miscellaneous proposals (see some instances in Figure \ref{fig:block17}).

\noindent To have more insight on the impact of the number of orientations in terms of performance and efficiency, we computed,  in the second experiment, the $\mu_{AP}$ and the average processing time per image, 
for different values of $K$ ranging from 2  to 8 while keeping the number of selected tensors \textit{M} fixed to 2. In the results, reported in Table \ref{tab:ablationK},  we can notice that the detection performance slightly increases with the number of orientations in both datasets. However, this little enhancement comes at the detriment of the computational time, as reflected in the observed exponential increase rate in both datasets.  For instance,  the computation time increases by a factor of two and four, when the number of orientations passes from $K=4$ to $K=5$, for the GDXray and the SIXray, respectively. These observations indicate that the volume of generated proposals become excessively redundant after the number of orientation goes beyond a certain threshold.  Considering the figures in Table \ref{tab:ablationKM} and  Table \ref{tab:ablationK}, we choose  $K=4$ and $M=2$ as the optimal values for the number of orientations and the number of tensors, as they present the best trade-off between the detection performance and the computational time.
\begin{table}[htb]
\centering
    \caption{Detection performance in terms of $\mu_{AP}$ by varying number of  orientations and the number of selected tensors for the GDXray and the SIXray.}
\begin{minipage}{.49\linewidth}
\resizebox{.9\linewidth}{!}{
\begin{tabular}{lc|lll}
\multicolumn{2}{l|}{\multirow{2}{*}{GDXray}} & \multicolumn{3}{c}{K}                                                \\
\multicolumn{2}{l|}{}                        & \multicolumn{1}{c}{2} & \multicolumn{1}{c}{3} & \multicolumn{1}{c}{4} \\ \hline
    \toprule
\multirow{10}{*}{M} & \multicolumn{1}{c|}{1}  & 0.593                  & 0.652                  & 0.798                 \\
                    & \multicolumn{1}{c|}{2}  & 0.671                  & 0.774                  & \textbf{0.934}               \\
                    & \multicolumn{1}{c|}{3}  & 0.764                  & 0.836                  & 0.913                 \\
                    & \multicolumn{1}{c|}{4}  & -                     & 0.769                  & 0.879                 \\
                    & \multicolumn{1}{c|}{5}  & -                     & 0.684                  & 0.742                 \\
                    & \multicolumn{1}{c|}{6}  & -                     & 0.543                  & 0.617                 \\
                    & \multicolumn{1}{c|}{7}  & -                     & -                     & 0.536                 \\
                    & \multicolumn{1}{c|}{8}  & -                     & -                     & 0.361                 \\
                    & \multicolumn{1}{c|}{9}  & -                     & -                     & 0.295                 \\
                    & \multicolumn{1}{c|}{10} & -                     & -                     & 0.225                
\end{tabular}
}
\end{minipage}
\begin{minipage}{.49\linewidth}
\resizebox{.9\linewidth}{!}{
\begin{tabular}{lc|lll}
\multicolumn{2}{l|}{\multirow{2}{*}{SIXray}} & \multicolumn{3}{c}{K}                                                \\
\multicolumn{2}{l|}{}                        & \multicolumn{1}{c}{2} & \multicolumn{1}{c}{3} & \multicolumn{1}{c}{4} \\ \hline
    \toprule
\multirow{10}{*}{M}           & 1           & 0.604                  & 0.738                  & 0.807                 \\
                              & 2           & 0.749                  & 0.826                  & \textbf{0.959}              \\
                              & 3           & 0.801                  & 0.804                  & 0.894                 \\
                              & 4           & -                     & 0.743                  & 0.890                 \\
                              & 5           & -                     & 0.662                  & 0.697                 \\
                              & 6           & -                     & 0.581                  & 0.573                 \\
                              & 7           & -                     & -                     & 0.498                 \\
                              & 8           & -                     & -                     & 0.396                 \\
                              & 9           & -                     & -                     & 0.216                 \\
                              & 10          & -                     & -                     & 0.164                
\end{tabular}
}
\end{minipage}
\label{tab:ablationKM}
\end{table}
\begin{table}[htb]
    \centering
    \caption{Performance in terms of $\mu_{AP}$ and detection time by varying the number of orientations \textit{K}. The number of selected tensors \textit{M} is set to 2 here. Time is shown in seconds for processing one image on average.}
    \begin{tabular}{c c c c c}
        \toprule
         \multirow{2}{1em}{$K$} &  \multicolumn{2}{c}{GDXray} &  \multicolumn{2}{c}{SIXray}\\
         \cline{2-5}
         & $\mu_{AP}$ & Time (sec) & $\mu_{AP}$ & Time (sec) \\
         \hline
            2	& 0.9212	& 0.0051	& 0.9486	& 0.0093 \\
            3	& 0.9285	& 0.0103	& 0.9532	& 0.0165 \\
            \textbf{4}	& \textbf{0.9343}	& \textbf{0.0174}    & \textbf{0.9595}	& \textbf{0.0212} \\
            5	& 0.9386	& 0.0463	& 0.9617	& 0.3879 \\
            6	& 0.9408	& 0.6540	& 0.9643	& 1.5001 \\
            7	& 0.9411	& 1.1793	& 0.9691	& 2.0791 \\
            8	& 0.9437	& 2.6213	& 0.9738	& 2.5457 \\
         \toprule
    \end{tabular}
    \label{tab:ablationK}
\end{table}

\noindent To analyze the performance of the CST framework across different pre-trained models, we conducted another series of experiments, in which we computed the  $\mu_{AP}$ for several standard classification networks.  The results, depicted in Table \ref{tab:ablation3}, show that the best detection performance is achieved by DenseNet\textsubscript{201}, whereas ResNet\textsubscript{101} stood second and ResNet\textsubscript{50} came third. 
However the  DenseNet\textsubscript{201} is outperforming ResNet\textsubscript{50} only by 0.722\% on GDXray and 0.08\% on SIXray dataset. Also, the ResNet models have better trade-offs between memory consumptions and time performance than DenseNets \cite{_67}, due to which, we prefer the ResNets (particularly ResNet\textsubscript{50} because of its less memory consumption) in the proposed study.
\begin{table}[t]
    \centering
    \caption{Comparison of pre-trained models for baggage items proposals recognition in terms of $\mu_{AP}$. Bold indicates the best results. Second best results are underlined while optimal results are highlighted in blue}
    \begin{tabular}{c c c}
        \toprule
         \textbf{Model} &  \textbf{GDXray} & \textbf{SIXray} \\
        \hline
            VGG\textsubscript{16} + CST  &	 0.8936  &	 0.9126 \\
            \color{blue}ResNet\textsubscript{50} + CST  &	 \color{blue}0.9343  &	 \color{blue}0.9595 \\
            \underline{ResNet\textsubscript{101} + CST}  &	 \underline{0.9401}  &	 \underline{0.9564} \\
            GoogleNet + CST  &	 0.9296  &	 0.9419 \\
            \textbf{DenseNet\textsubscript{201} + CST}  &	 \textbf{0.9411}  &	 \textbf{0.9603} \\
         \toprule
    \end{tabular}
    \label{tab:ablation3}
\end{table}

\begin{table}[htb]
    \centering
    \caption{Mean $IoU$ ratings on GDXray and SIXray for extracting different suspicious items}
    \begin{tabular}{c c c}
        \toprule
         \textbf{Items} &  \textbf{GDXray} & \textbf{SIXray} \\
        \hline
         Gun & 0.9487* &	0.9811 \\
         Knife & 0.9872 &	0.9981 \\
         Shuriken & 0.9658 &	- \\
         Chip & 0.9743 &	- \\
         Mobile & 0.9425 &	- \\
         Razor & 0.9681 &	- \\
         Wrench & -	& 0.9894 \\
         Pliers & -	& 0.9637 \\
         Scissor & - &	0.9458 \\
         Hammer & -	& 0.9354 \\
         \hline
         \textbf{Mean $\pm$ STD} & \textbf{0.9644 $\pm$ 0.0165} & \textbf{0.9689 $\pm$ 0.0249} \\
         \toprule
    \end{tabular}
    \begin{tablenotes}
        \item[*] * the $IoU$ score of 'Gun' category in GDXray is an average of \textit{pistols} and \textit{revolvers}.
    \end{tablenotes}
    \label{tab:my_label2}
\end{table}

\subsection{System Validation}
\label{system-validation}
\begin{figure}[htb]
\begin{center}
\includegraphics[scale=0.076]{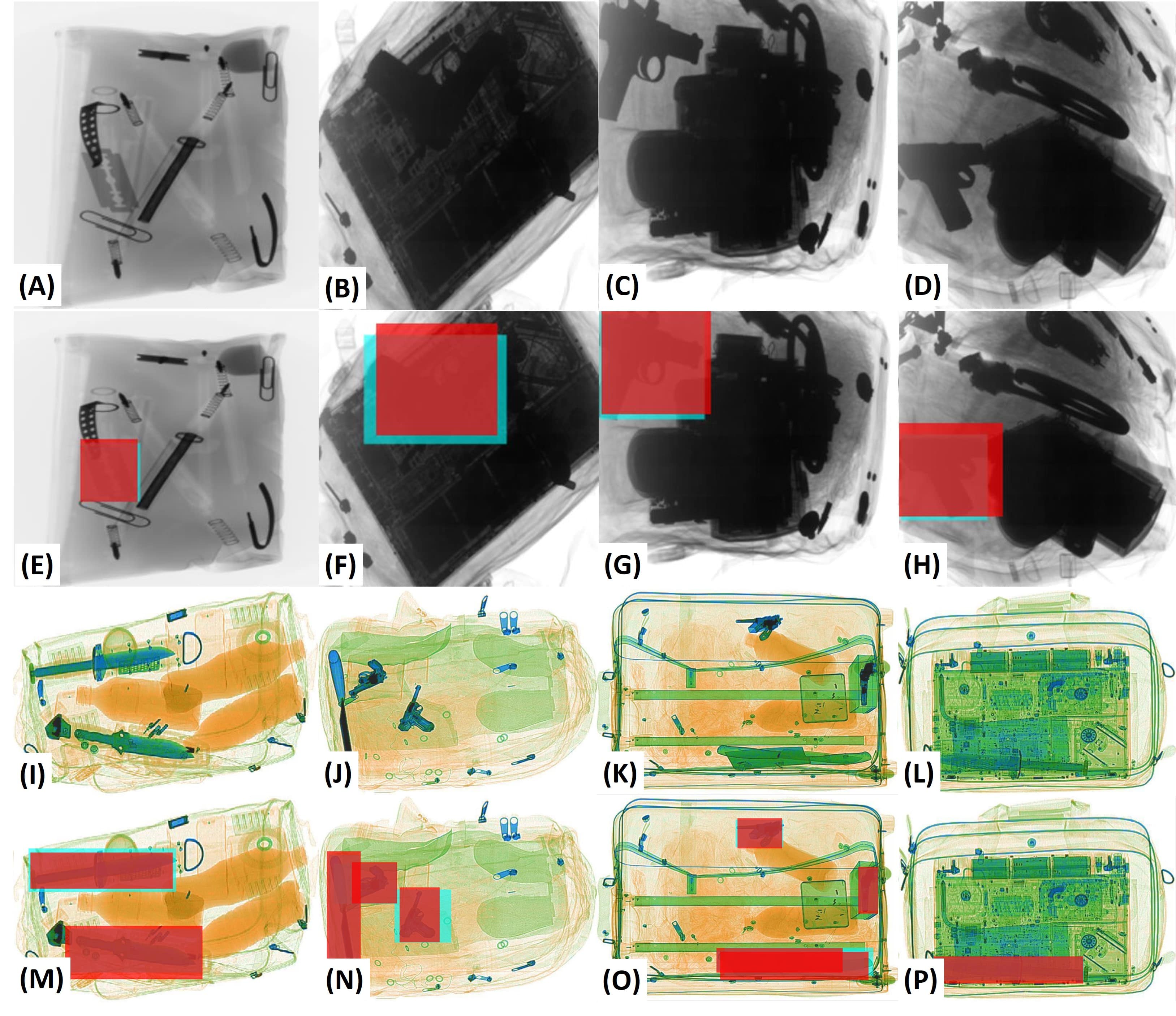}
\end{center}
\caption{\small Examples of detected objects for the GDXray and the SIXray  dataset.  Red color shows the extracted regions, while cyan color shows the ground truth. These examples showcase the extraction accuracy of our framework. }
\centering
\label{fig:block11}
\end{figure}

\noindent{In Figure \ref{fig:block11}, we report some qualitative results for the two datasets, 
showcasing how our framework can effectively extract and localize suspicious items from the grayscale and color X-ray scans w.r.t their ground truths.}

\begin{figure*}[htb]
\begin{center}
\includegraphics[scale=0.107]{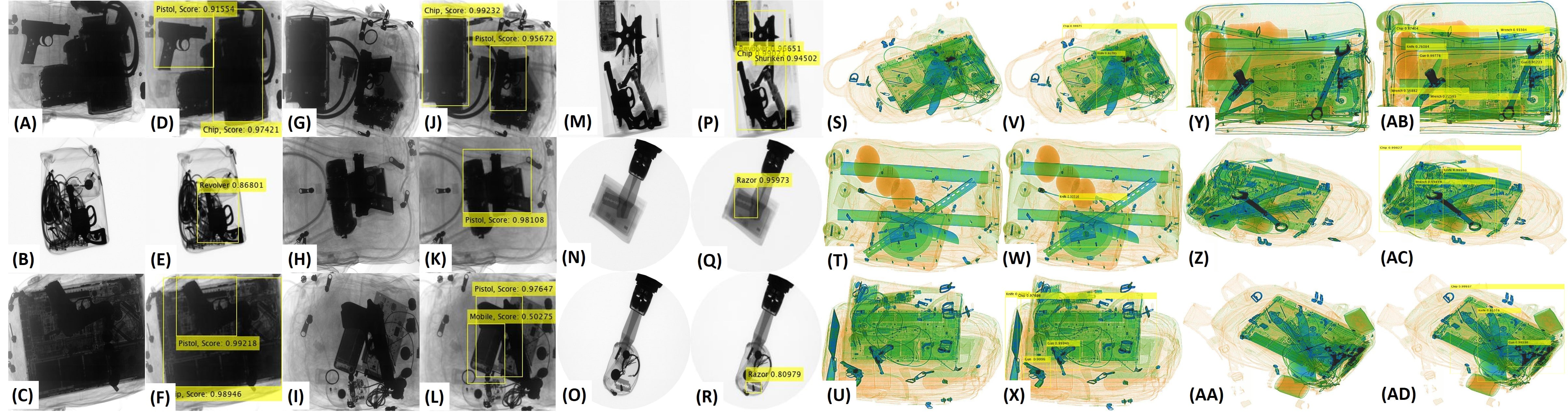}
\end{center}
\caption{\small Examples showcasing the performance of the proposed framework in recognizing heavily occluded, cluttered and concealed items from GDXray and the SIXray. The first, third, fifth, seventh and ninth column shows the original scans. Please zoom for better visibility.  }
\centering
\label{fig:block13}
\end{figure*}

\noindent In Figure \ref{fig:block13}, we report other qualitative results of items exhibiting high occlusion and clutter. Through these examples, we can appreciate the capacity of the framework for accurately extracting and recognizing items in such challenging conditions. For example, in (A, D) and (B, E), we can observe that the \textit{chip} and  \textit{revolver} have been detected despite being severely occluded by the bundle of wires. Similar examples can be also noticed for the occluded \textit{pistol}, \textit{mobile}, \textit{shuriken} and \textit{razor blades} in  (C, F), (G, J), (H, K), (I, L), (M, P), (N, Q) and (O, R). In Q and R, in particular, we can observe how effectively the proposed framework has identified the \textit{razor blade} despite having and intensity close to the background. 

\noindent For the SIXray dataset, we can also observe how our framework effectively recognizes partially and heavily occluded suspicious items (see the pairs (S, V), (T, W), (U, X), (Y, AB), (Z, AC), and (AA, AD)).  Notice, in particular, the detected overlapped \textit{knife} in (S, V) and (T, W) and the overlapped \textit{wrench} and \textit{knife} in (Y, AB) and (Z, AC). 
Also, these examples reflect the effectiveness of the coherent tensor framework in discriminating overlapped items even when there are a lot of background edges and sharp transitions within the scan as it can be seen in (S, V), (T, W), (U, X), (Z, AC), and (AA, AD). Also, the ability of the proposed framework to discriminate between \textit{revolver} and \textit{pistol} can be in noticed in (B, E) for the \textit{revolver} and
((C, F), (I, L)) for the \textit{pistol}.   Note that all the existing state-of-the-art solutions have considered these two items as part of the single  \textit{handgun}  category in their methods.  

\begin{figure*}[t]
\begin{center}

\includegraphics[scale=0.406]{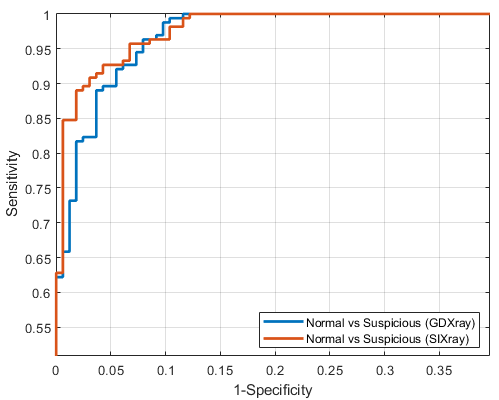}
\includegraphics[scale=0.406]{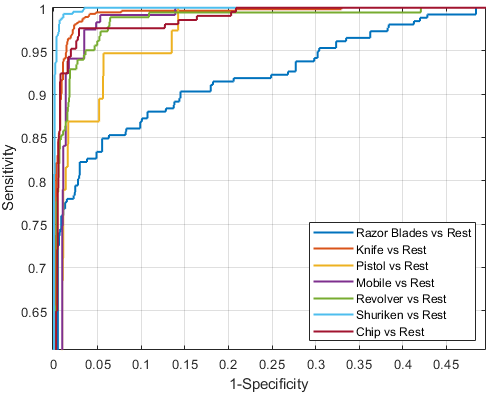}
\includegraphics[scale=0.406]{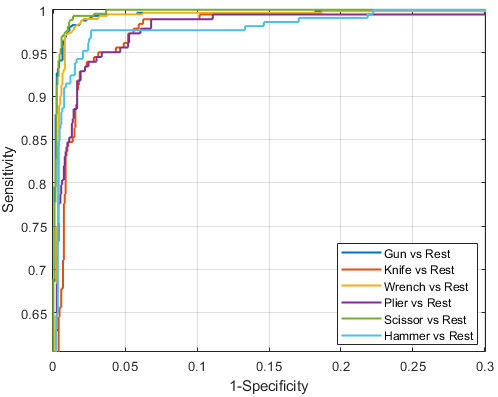}



\end{center}
\caption{\small From left to right: ROC curves for identifying normal and suspicious items on GDXray and SIXray dataset, ROC curve for items recognition on GDXray and SIXray dataset, respectively.
}
\centering
\label{fig:block15}
\end{figure*}

\noindent Figure \ref{fig:block15} reports the validation of our system through the ROC curves. Figure \ref{fig:block15} (left) depicts the recognition rate related to the classification normal versus suspicious items, for the GDXray and the SIXray datasets. We achieved here the $AUC$ score of 0.9863 and 0.9901, respectively. 
Figure \ref{fig:block15}  (middle) and (right) show the ROC curves related to the item-wise recognition for the  GDXray and SIXray, respectively.
Here the true positives and true negatives represent pixels of the items and pixels of the background which are correctly identified, respectively. We note here that the reported scores are computed based on both:   the correct extraction and the correct recognition of the suspicious items. For example, if the item has been correctly extracted by the CST framework, but it has been misclassified by the ResNet\textsubscript{50}, then we counted it as a false negative in the scoring.
In Figure \ref{fig:block15} (middle), we notice the  \textit{razor blades} score relatively lower than the other items (This is also reflected in the related  $AUC$ score of 0.9582 in Table \ref{tab:auc}). This is explained by the fact that the intensity differences between the \textit{razor blades} and the background is very minimum within the scans of the GDXray dataset causing this item to be missed by the CST at some instances.   

\begin{figure}[htb]
\begin{center}

\includegraphics[scale=0.337]{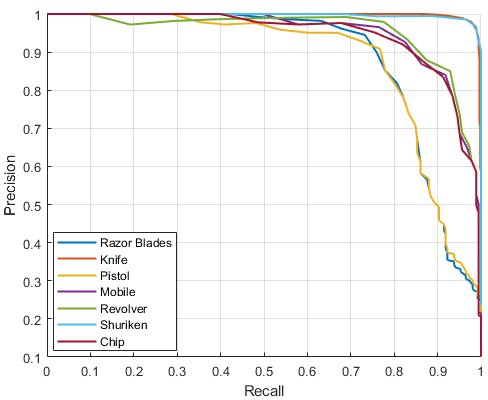}
\includegraphics[scale=0.337]{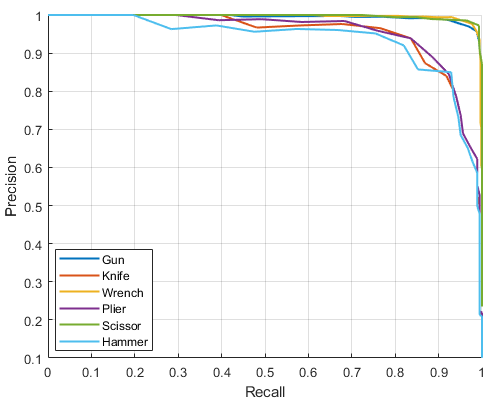}

\end{center}
\caption{\small From left to right: $P_{RC}$ curves for items recognition on GDXray and SIXray dataset.
}
\centering
\label{fig:block15N}
\end{figure}

 \noindent Figure \ref{fig:block15N} shows the $P_{RC}$ curves computed on the GDXray and SIXray dataset, respectively.  The $P_{RC}$ confirms further the robustness of our framework.

 \noindent In Table \ref{tab:mu}, we report the $\lambda_P$ performance for each item in the two datasets. On average, our system achieves a $\mu_{AP}$ score of 0.9343 and 0.9595 on GDXray and SIXray datasets, respectively. Note that the $\lambda_P$ score of 0.9101 for the \textit{handgun} category in the GDXray dataset is the average of \textit{pistol} and \textit{revolver} scores. 
 
 \noindent Table \ref{tab:auc} depicts the $AUC$ scores for the different items, with mean scores of 0.9878 and 0.9950 obtained for the GDXray and the SIXray dataset, respectively. As in Table \ref{tab:mu}, the $AUC$ score of the \textit{Gun} class is the average of the \textit{pistol} and the \textit{revolver} scores.
\begin{table}[t]
    \centering
    \caption{$\lambda_P$ and $\mu_{AP}$ scores on GDXray and SIXray for the detection of different suspicious items. '-' indicates that the respective item is not present in the dataset}
    \begin{tabular}{c c c}
        \toprule
         \textbf{Items} &  \textbf{GDXray} & \textbf{SIXray} \\
        \hline
            Razor Blades & 0.8826 &	- \\
            Knife & 0.9945 &	0.9347 \\
            Pistol & 0.8762	& - \\
            Mobile & 0.9357	& - \\
            Revolver & 0.9441 &	- \\
            Shuriken & 0.9917 &	- \\
            Chip & 0.9398 &	- \\
            Gun & 0.9101* & 0.9911 \\
            Wrench & - & 0.9915 \\
            Plier & - &	0.9267 \\
            Scissor & -	& 0.9938 \\
            Hammer & -	& 0.9189 \\
         \hline
         \textbf{Mean $\pm$ STD} & \textbf{0.9343  $\pm$ 0.0442} & \textbf{0.9595 $\pm$ 0.0362} \\
         \toprule
    \end{tabular}
    \begin{tablenotes}
         \item[*] * the $\lambda_P$ score of 'Gun' category in GDXray is an average of \textit{pistols} and \textit{revolvers}.
    \end{tablenotes}
    \label{tab:mu}
\end{table}
\begin{table}[b]
    \centering
    \caption{$AUC$ scores on GDXray and SIXray for the detection of different suspicious items. '-' indicates that the respective item is not present in the dataset}
    \begin{tabular}{c c c}
        \toprule
         \textbf{Items} &  \textbf{GDXray} & \textbf{SIXray} \\
        \hline
            Razor Blades & 0.9582 &	- \\
            Knife & 0.9972 &	0.9981 \\
            Pistol & 0.9834	& - \\
            Mobile & 0.9914	& - \\
            Revolver & 0.9932 &	- \\
            Shuriken & 0.9987 &	- \\
            Chip & 0.9925 &	- \\
            Gun & 0.9883* & 0.9910 \\
            Wrench & - & 0.9971 \\
            Plier & - &	0.9917 \\
            Scissor & -	& 0.9990 \\
            Hammer & -	& 0.9932 \\
         \hline
         \textbf{Mean $\pm$ STD} & \textbf{0.9878  $\pm$ 0.0120} & \textbf{0.9950 $\pm$ 0.0031} \\
         \toprule
    \end{tabular}
    \begin{tablenotes}
         \item[*] * the $AUC$ score of 'Gun' category in GDXray is an average of \textit{pistols} and \textit{revolvers}.
    \end{tablenotes}
    \label{tab:auc}
\end{table}
\subsection{Comparative Study}
\label{comparative-study}
\noindent For the GDXray dataset, we compared our framework with the methods \cite{_39}, \cite{_42}, \cite{_57} and \cite{_58} as shown in Table \ref{tab:comp1}. The performance comparison is nevertheless indirect as the experiment protocol in each study differs. We, (as well as authors in \cite{_39}) followed the standards laid in \cite{_55} by considering 400 images for training (100 for \textit{razor blades}, 100 for \textit{shuriken} and 200 for \textit{handguns}.  However, \cite{_39} used 600 images for testing purposes (200 for each item) and considered only 3 items whereas we considered 7 items and used 7,362 scans for testing.  The authors in \cite{_42} considered a total of 3,669 selective images in their work having 1,329 \textit{razor blades}, 822 \textit{guns}, 540 \textit{knives}, and 978 \textit{shuriken}. To train Faster R-CNN, YOLOv2 and Tiny YOLO models, they picked 1,223 images from the dataset and augmented them to generate 2,446 more images. The work reported in \cite{_58} involved 18 images only while \cite{_57} reports a study that is based on non-ML methods where the authors conducted 130 experiments to detect \textit{razor blades} within the X-ray scans.

\noindent Contrary to the aforementioned methods, we assessed our framework for all the performance criteria described in Section \ref{sec:expsetuo} (C). Also, it should be noted here that the proposed framework has been evaluated in the most restrictive conditions as compared to its competitors where the true positive samples (of the extracted items) were only counted towards the scoring when they were correctly classified by the ResNet\textsubscript{50} model as well. So, if the item has been correctly extracted by the CST framework but was not correctly recognized by the ResNet\textsubscript{50} model, we counted it as a misclassification in the  evaluation. Despite these strict 
requirements,  we were able to achieve 2.37\% improvements in terms of precision as evidenced in Table \ref{tab:comp1}. 

\noindent For the SIXray dataset, we compared our system with the methods proposed in \cite{_46} and \cite{_32} (the only two frameworks which have been applied on the SIXray dataset to date). Results are reported in Table \ref{tab:comp2}.
The SIXray dataset is divided into three subsets to address the problem of class imbalance. These subsets are named as SIXray10, SIXray100 and SIXray1000. SIXray10 contains all 8,929 positive scans (having suspicious items) and 10 times the negative scans (which do not contain any suspicious item). Similarly, SIXray100 has all the positive scans and 100 times the negative scans. SIXray1000 contains only 1000 positive scans and all the negative scans (1,050,302 in total). So, the most challenging subset for the class imbalance problem is SIXray1000. Note that  the  works \cite{_46} and \cite{_32} employed different pre-trained models, which we also reported in Table \ref{tab:comp2} for completeness. 
Moreover, for a direct and fair comparison with \cite{_46} and \cite{_32}, we have trained the proposed framework on each subset of the SIXray dataset individually and evaluated it using the same metrics as described in \cite{_46}. Furthermore, we have excluded the \textit{hammer} class in these experiments as it was not considered in \cite{_46} and \cite{_32}. The scores depicted in Table \ref{tab:comp2} evidence the superiority of our framework over its competitors in terms of object classification and localization. 
\noindent In the last experiment, we compared the computational performance of our system with standard one-staged (such as YOLOv3 \cite{yolov3}, YOLOv2 \cite{_62} and RetinaNet \cite{_11}), and the two-staged detectors (such as Faster R-CNN \cite{_63}), as they have been widely used for suspicious items detection in the literature. 
The results, depicted in Table \ref{tab:time}, show that our system scores the best average time performance in both training and testing outperforming the existing object detectors. Note also that although YOLOv2 \cite{_62}  produces significantly improved computational performance over other two-staged architectures, it has limited capacity for detecting smaller objects like \textit{diapers}, \textit{sunglasses}, and \textit{rubber erasers} \cite{_62}. Though  YOLOv3 \cite{yolov3} seems improving in this regard. 

\begin{table*}[htb]
\footnotesize
  \centering
  \caption{Performance comparison of the proposed framework on the GDXray dataset. The first and second-best performances are marked in bold and blue, respectively. '-' indicates that the metric is not computed. The protocols are defined below.}
  \begin{center}
    \begin{tabular}
    { m{1.5cm}  m{1.5cm}   m{1.5cm}   m{1.1cm}  m{1.5cm}  m{0.8cm}  m{0.8cm}  m{0.8cm}  m{0.8cm}  m{0.8cm}  m{0.7cm}  m{0.6cm} }
	\toprule
	\textbf{Metric}  &	\textbf{Proposed} &		\textbf{Faster R-CNN \cite{_42}} &	\textbf{YOLOv2 \cite{_42}} &	\textbf{Tiny YOLO \cite{_42}} &	\textbf{AISM\textsubscript{1} \cite{_39}\tnote{*}}	 &	\textbf{AISM\textsubscript{2} \cite{_39}\tnote{*}}	 &	\textbf{SURF \cite{_39}}	 & \textbf{SIFT \cite{_39}}	 &		\textbf{ISM \cite{_39}}	 &	\textbf{\cite{_57}} &	\textbf{\cite{_58}} \\
	\hline
	AUC & \color{blue}{0.9870}  &	- &	- &		- &	\textbf{0.9917} &  \textbf{0.9917} &		0.6162 &	0.9211	 &	0.9553 &	- &	- \\
	\hline
	Accuracy &	0.9683 & \textbf{0.9840}	& \color{blue}{0.9710} &	0.89 &	- &	- &	- &	- &	- &	- &	- \\
    \hline
    T\textsubscript{PR} &	0.8856	&	0.98	&	0.88	&	0.82	&	\textbf{0.9975}		& \color{blue}{0.9849}	&	0.6564	&	0.8840	&	0.9237	&	0.89	&	0.943 \\
    \hline
    Specificity	 &	\textbf{0.9890}	&	-	&	-	&	-	&	0.95	&	\color{blue}{0.9650}	&	0.63	&	0.83	&	0.885	&	-	&	0.944 \\
    \hline
    F\textsubscript{PR} &	\textbf{0.0110}	&	-	&	-	&	-	&	0.05	&	\color{blue}{0.035}	&	0.37	&	0.17	&	0.115	&	-	&	0.056 \\
    \hline
    P\textsubscript{R}  &	\textbf{0.9526}	&	\color{blue}{0.93}	&	0.92	&	0.69	&	-	&	-	&	-	&	-	&	-	&	0.92	&	- \\
    \hline
     F\textsubscript{1}  & \color{blue}{0.9178} &	\textbf{0.9543} &	0.8996 &	0.7494 &	- &	- &	- &	- &	- &	0.9048 &	- \\
    \toprule
    \end{tabular}
    \end{center}
    \begin{tablenotes}[flushleft]
            \item[a]   Proposed: Classes: 7, Split: 5\% for training and 95\% for testing, Training Images: 400 (and 388 more for extra items), Testing Images: 7,362.
            
            \item[b] \cite{_39}: Classes: 3, Split: 40\% for training, 60\% for testing, Training Images: 400, Testing Images: 600 (200 for each category).
            
            \item[c] \cite{_42}: Classes: 4, Split: 80\% for training and 20\% for validation, Total Images: 3,669.
            
            \item[d] \cite{_57}: Classes: 1, (non-ML approach).
            
            \item[e] \cite{_58}: Classes: 3, Total Images: 18.
            \item[f]   * the ratings of AISM\textsubscript{1} and AISM\textsubscript{2} are obtained from the ROC curve of AISM for different $F_{PR}$ values.
    \end{tablenotes}
\label{tab:res}

\vspace{-0.2cm}
\label{tab:comp1}
\end{table*}

\begin{table*}[htb]
\footnotesize
  \centering
  \caption{Performance comparison of the proposed framework with existing solutions on the SIXray subsets. The first and second-best performances are marked in bold and blue, respectively.}

\begin{tabular}{ m{1.5cm}  m{1.5cm}    m{1cm}   m{1.0cm}   m{1.4cm}  m{1.5cm}  m{1.4cm}  m{1.5cm}  m{1.5cm}  m{0.8cm}  }
	\toprule
    \textbf{Criteria} &	\textbf{Subset} &	\textbf{ResNet\textsubscript{50} + CST} &	\textbf{ResNet\textsubscript{50} \cite{_48}} &	\textbf{ResNet\textsubscript{50} + CHR \cite{_46}}	 & \textbf{DenseNet \cite{_59}} &	\textbf{DenseNet + CHR \cite{_46}}	& \textbf{Inceptionv3 \cite{_60}}	& \textbf{Inceptionv3 + CHR \cite{_46}} &	\textbf{\cite{_32}} \\
	\hline
	 \multirow{3}{4em}{Mean Average Precision} & SIXray10 &	\textbf{0.9634} &	0.7685 &	0.7794 &	0.7736 &	0.7956 &	0.7956 &	0.7949 &	\color{blue}{0.86} \\
	  &  SIXray100 &	\textbf{0.9318} &	0.5222 &	0.5787 &	0.5715 &	\color{blue}{0.5992} &	0.5609 &	0.5815 &	- \\
    & 	SIXray1000 &	\textbf{0.8903} &	0.3390 &	0.3700 &	0.3928 &    \color{blue}{0.4836} &	0.3867 &	0.4689 &	- \\
    \hline
    \multirow{3}{4em}{Localization Accuracy} & SIXray10 &	\textbf{0.8413} &	0.5140 &	0.5485 & 	0.6246 & 	\color{blue}{0.6562} &	0.6292 &	0.6354 &	- \\
 	& SIXray100	& \textbf{0.7921} &	0.3405 &	0.4267 & 	0.4470 &	\color{blue}{0.5031} &	0.4591 	& 0.4953 &	- \\
  	& 	SIXray1000  &	\textbf{0.7516}  &	0.2669 &	0.3102 &	0.3461 &	\color{blue}{0.4387} &	0.3026 &	0.3149 &	- \\
\toprule
\end{tabular}
\label{tab:comp2}

\vspace{-0.2cm}
\end{table*}

\begin{table}[htb]
\footnotesize
\center
\caption{Time performance comparison of the proposed framework with popular object detectors.}
\begin{tabular}{ccc}
\toprule 
Frameworks & Training (sec) & Testing (sec) \\ \toprule
YOLOv3 \cite{yolov3}      & 694.81   & 0.021   \\ \hline
YOLOv2 \cite{_62}      & 712.06   & 0.024   \\ \hline
RetinaNet \cite{_11}    & 926.97   & 0.071   \\ \hline
R-CNN \cite{_8}        & 305,896   & 134.13  \\ \hline
Faster R-CNN \cite{_63} & 19,359    & 0.52    \\ \hline
\textbf{Proposed}     & \textbf{679.43}  & \textbf{0.020}  \\ \toprule
\end{tabular}
\label{tab:time}
\end{table}

\begin{figure}
\begin{center}
\includegraphics[scale=0.075]{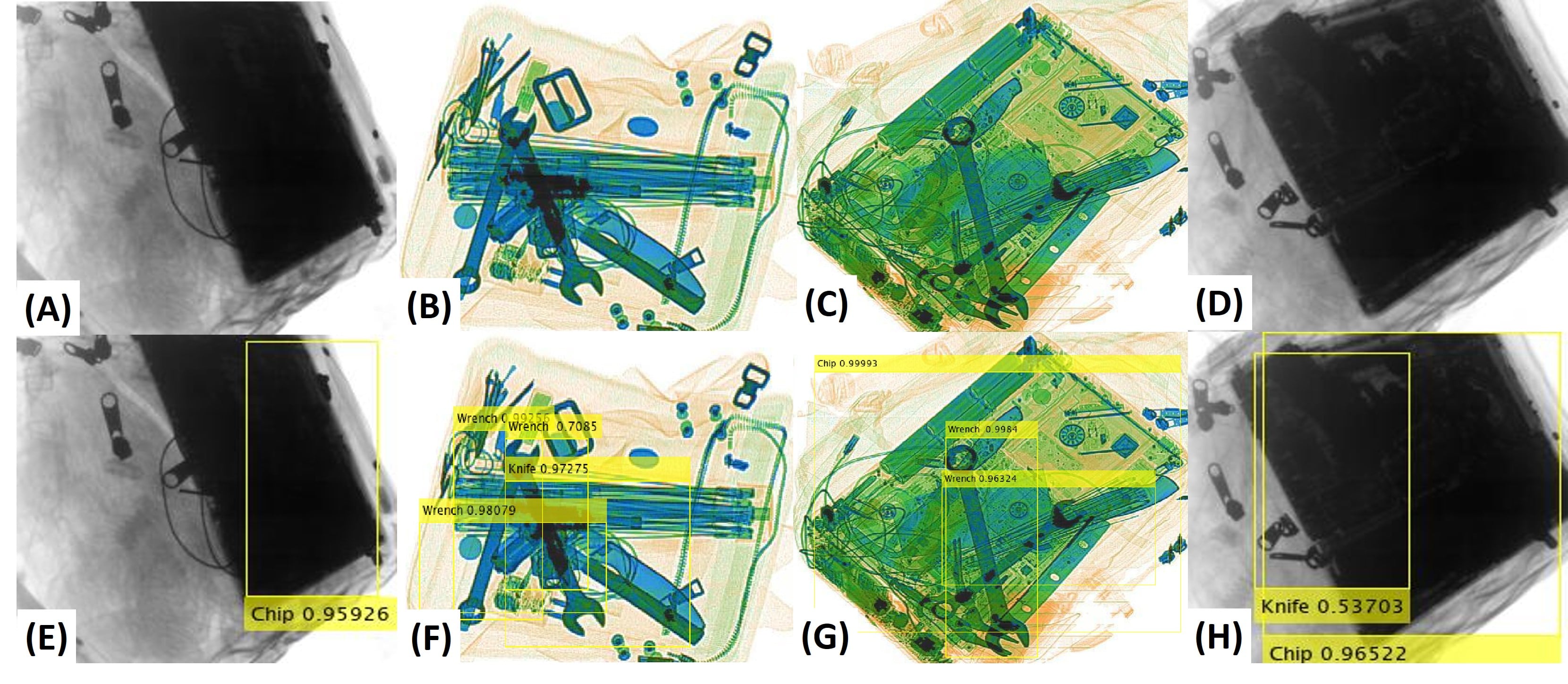}
\end{center}
\caption{\small Failure cases on GDXray and SIXray datasets. Left column shows the original scans while right column shows the results. }
\centering
\label{fig:block18}
\end{figure}
\subsection{Failure Cases}
\label{failure-cases}
 \noindent We observed two types of scenarios where our framework is a bit limited in extracting the suspicious items correctly. The first scenario is when the CST framework cannot highlight the transitional differences in the extremely occluded objects. One of such cases is shown in Figure \ref{fig:block18} (A, E), where we can see that our framework could not  identify the \textit{pistol} due to the complex grayscale nature of the scan. However, identifying a \textit{pistol} in such a challenging scan is an extremely difficult task for even a human expert. Also, a \textit{gun} item was not detected in (B, F) because of the extreme clutter in that scan. In (C, G), two instances of a \textit{knife} could not be detected. While these limitations have been rarely observed in the  experiments  they can be catered by considering more orientations to further reveal the transitional patterns. However, doing so would increase the overall computational time of the proposed framework. 

\noindent The second scenario  is related to cases of a correct proposal generation with misclassification of their content. An instance of these cases is depicted in 
Figure \ref{fig:block18} (H) (last column), in which the \textit{pistol} object has been misclassified as \textit{knife}.   However,  we did notice that such instances are detected with a relatively low score (e.g. 0.537 for that \textit{knife} item). These cases can, therefore, be  catered in the second screening stage based on their confidence score.  Note  that detecting suspicious items even misclassified  is a safer approach in this context.

\noindent In some instances, we also observed that our framework sometimes does not generate tight bounding boxes  for the recognized items such as the \textit{chip} in Figure \ref{fig:block13} (Y, AB), (Z, AC); and the \textit{razor blade} in (N, Q). This limitation emanates from our contour-based proposal identification in the CST framework in which the bounding boxes are  not necessarily tight to the object. While bounding box regression can address this limitation, such a mitigation approach will incur a significant additional computational burden in return of  a marginal impact on the accuracy.
\section{Conclusion} \label{sec:conc}
\noindent This paper presents a deep learning system for the identification of heavily cluttered and occluded suspicious items from X-ray images. In this system, we proposed an original contour-based proposal generation using a Cascaded Structure Tensor (CST) framework.  
The proposed framework is highly sensitive in picking merged, cluttered and overlapping items at a different level of intensity through an original iterative proposal generation scheme.  
The system recognizes the proposals using single feed-forward CNN architecture and does not require any exhaustive searches and regression networks. This property makes our system more time-efficient as compared to the popular CNN object detectors.  
The proposed system is rigorously tested on different publicly available datasets and is thoroughly compared with existing state-of-the-art solutions using different metrics.  We achieve the mean $IoU$ score of 0.9644 and 0.9689, $AUC$ score of 0.9878 and 0.9950, and a $\mu_{AP}$ score of 0.9343 and 0.9595 on GDXray and SIXray dataset, respectively.
The proposed framework outperforms the state-of-the-art in terms of quantitative and computational performance. The proposed system can be extended to normal photographs and popular large scale publicly available datasets for the automatic object detection. This will be the object of our future research.

\section*{Acknowledgment}
\noindent{This work is supported with a research fund from Khalifa University: Ref: CIRA-2019-047.} 

\small

\end{document}